%% file: main.tex
\documentclass{article}

\PassOptionsToPackage{square,sort&compress,numbers}{natbib}

\usepackage[final]{neurips_2024}

\usepackage{float}
\usepackage{wrapfig}
\usepackage{hyperref}       
\hypersetup{
    colorlinks=true,
	linkcolor=blue,
	filecolor=magenta,      
	urlcolor=blue,
	citecolor=blue,
	pdfinfo={
        Title={Humanity's Last Exam},
        Subject={machine learning, large language models, datasets and benchmarks},
        Keywords={machine learning, large language models, datasets and benchmarks},
    }
}

\usepackage[utf8]{inputenc} 
\usepackage[T1]{fontenc}    
\usepackage{hyperref}       
\usepackage{url}            
\usepackage{booktabs}       
\usepackage{amsfonts}       
\usepackage{nicefrac}       
\usepackage{microtype}      
\usepackage{xcolor}         
\usepackage{comment}
\usepackage{multicol}
\usepackage{svg}

\usepackage{graphicx}

\usepackage{amsmath}
\usepackage{xcolor}
\usepackage{soul}
\usepackage{cleveref}
\usepackage{listings}
\usepackage{tikz}
\usepackage{eso-pic}

\let\svthefootnote\thefootnote
\newcommand\freefootnote[1]{%
  \let\thefootnote\relax%
  \footnotetext{#1}%
  \let\thefootnote\svthefootnote%
}

\usepackage{xcolor}
\usepackage{colortbl}
\usepackage{tcolorbox}

\definecolor{darkred}{RGB}{255,150,150}
\definecolor{lightred}{RGB}{255,200,200}
\definecolor{lightorange}{RGB}{255,230,200}
\definecolor{lightyellow}{RGB}{255,255,200}
\definecolor{lightgreen}{RGB}{220,255,220}
\definecolor{lightblue}{RGB}{200,220,255}
\definecolor{lightpurple}{RGB}{230,190,255}
\definecolor{lightteal}{RGB}{200,255,255}

\clearpage
\title{Humanity's Last Exam}

\begin{document}

\newcommand{\fullname}{\textsc{Humanity's Last Exam}}
\newcommand{\name}{\textsc{HLE}}
\newcommand{\questioncount}{$2{,}500$}

\newcommand{\claude}{\textsc{Claude 3.5 Sonnet}}
\newcommand{\geminiflash}{\textsc{Gemini 1.5 Flash}}
\newcommand{\geminipro}{\textsc{Gemini 1.5 Pro}}
\newcommand{\geminithinking}{\textsc{Gemini 2.0 Flash Thinking}}
\newcommand{\gptfouromini}{\textsc{GPT-4o mini}}
\newcommand{\gptfouro}{\textsc{GPT-4o}}
\newcommand{\oonemini}{\textsc{o1-mini}}
\newcommand{\oonepreview}{\textsc{o1-preview}}
\newcommand{\oone}{\textsc{o1}}
\newcommand{\groktwo}{\textsc{Grok 2}}
\newcommand{\othreemini}{\textsc{o3-mini}}
\newcommand{\deepseekrone}{\textsc{DeepSeek-R1}}
\newcommand{\othreeminimedium}{\textsc{o3-mini (medium)}}
\newcommand{\othreeminihigh}{\textsc{o3-mini (high)}}

\maketitle
\vspace{-50pt}

\subsection*{Organizing Team}
{\fontsize{8.5}{10}\selectfont
Long Phan$^{*1}$, Alice Gatti$^{*1}$, Ziwen Han$^{*2}$, Nathaniel Li$^{*2}$, 

Josephina Hu$^{2}$, Hugh Zhang$^\ddagger$, Chen Bo Calvin Zhang$^{2}$, Mohamed Shaaban$^{2}$, John Ling$^{2}$, Sean Shi$^{2}$, Michael Choi$^{2}$, Anish Agrawal$^{2}$, Arnav Chopra$^{2}$, Adam Khoja$^{1}$, Ryan Kim$^\dagger$, Richard Ren$^{1}$, Jason Hausenloy$^{1}$, Oliver Zhang$^{1}$, Mantas Mazeika$^{1}$, 

Summer Yue$^{**2}$, Alexandr Wang$^{**2}$, Dan Hendrycks$^{**1}$

}

{\small\textbf{$^{1}$ \textbf{Center for AI Safety}, $^{2}$ \textbf{Scale AI}}}

\freefootnote{\hspace{-1em}$^*$Co-first Authors. $^{**}$ Senior Authors. $^\dagger$ Work conducted while at Center for AI Safety. $^\ddagger$ Work conducted while at Scale AI. Complete list of author affiliations in \Cref{app:authors}. Correspondence to \url{agibenchmark@safe.ai}.}

\renewcommand{\thefootnote}{\fnsymbol{footnote}}
\subsection*{Dataset Contributors}
\input{sections/authors}

\renewcommand{\thefootnote}{\fnsymbol{footnote}}

\newpage
\input{sections/0-abstract}
\input{sections/1-introduction}

\input{sections/2-related-work}
\input{sections/3-dataset}

\input{sections/4-evaluation}
\input{sections/5-discussion}

\newpage

\bibliography{main}
\bibliographystyle{abbrvnat} 

\newpage
\appendix
\input{appendix/a-authors}
\input{appendix/b-dataset}
\input{appendix/c-evaluation}
\input{appendix/d-review_guidelines}
\end{document}

%% file: sections/authors.tex
{\fontsize{8.4}{10}\selectfont
Dmitry Dodonov, Tung Nguyen, Daron Anderson, Mikhail Doroshenko, Alun Cennyth Stokes, Mobeen Mahmood, Oleksandr Pokutnyi, Oleg Iskra, Jessica P. Wang, John-Clark Levin, Mstyslav Kazakov, Fiona Feng, Steven Y. Feng, Haoran Zhao, Michael Yu, Varun Gangal, Chelsea Zou, Zihan Wang, Serguei Popov, Robert Gerbicz, Geoff Galgon, Johannes Schmitt, Will Yeadon, Yongki Lee, Scott Sauers, Alvaro Sanchez, Fabian Giska, Marc Roth, Søren Riis, Saiteja Utpala, Noah Burns, Gashaw M. Goshu, Mohinder Maheshbhai Naiya, Chidozie Agu, Zachary Giboney, Antrell Cheatom, Francesco Fournier-Facio, Sarah-Jane Crowson, Lennart Finke, Zerui Cheng, Jennifer Zampese, Ryan G. Hoerr, Mark Nandor, Hyunwoo Park, Tim Gehrunger, Jiaqi Cai, Ben McCarty, Alexis C Garretson, Edwin Taylor, Damien Sileo, Qiuyu Ren, Usman Qazi, Lianghui Li, Jungbae Nam, John B. Wydallis, Pavel Arkhipov, Jack Wei Lun Shi, Aras Bacho, Chris G. Willcocks, Hangrui Cao, Sumeet Motwani, Emily de Oliveira Santos, Johannes Veith, Edward Vendrow, Doru Cojoc, Kengo Zenitani, Joshua Robinson, Longke Tang, Yuqi Li, Joshua Vendrow, Natanael Wildner Fraga, Vladyslav Kuchkin, Andrey Pupasov Maksimov, Pierre Marion, Denis Efremov, Jayson Lynch, Kaiqu Liang, Aleksandar Mikov, Andrew Gritsevskiy, Julien Guillod, Gözdenur Demir, Dakotah Martinez, Ben Pageler, Kevin Zhou, Saeed Soori, Ori Press, Henry Tang, Paolo Rissone, Sean R. Green, Lina Brüssel, Moon Twayana, Aymeric Dieuleveut, Joseph Marvin Imperial, Ameya Prabhu, Jinzhou Yang, Nick Crispino, Arun Rao, Dimitri Zvonkine, Gabriel Loiseau, Mikhail Kalinin, Marco Lukas, Ciprian Manolescu, Nate Stambaugh, Subrata Mishra, Tad Hogg, Carlo Bosio, Brian P Coppola, Julian Salazar, Jaehyeok Jin, Rafael Sayous, Stefan Ivanov, Philippe Schwaller, Shaipranesh Senthilkumar, Andres M Bran, Andres Algaba, Kelsey Van den Houte, Lynn Van Der Sypt, Brecht Verbeken, David Noever, Alexei Kopylov, Benjamin Myklebust, Bikun Li, Lisa Schut, Evgenii Zheltonozhskii, Qiaochu Yuan, Derek Lim, Richard Stanley, Tong Yang, John Maar, Julian Wykowski, Martí Oller, Anmol Sahu, Cesare Giulio Ardito, Yuzheng Hu, Ariel Ghislain Kemogne Kamdoum, Alvin Jin, Tobias Garcia Vilchis, Yuexuan Zu, Martin Lackner, James Koppel, Gongbo Sun, Daniil S. Antonenko, Steffi Chern, Bingchen Zhao, Pierrot Arsene, Joseph M Cavanagh, Daofeng Li, Jiawei Shen, Donato Crisostomi, Wenjin Zhang, Ali Dehghan, Sergey Ivanov, David Perrella, Nurdin Kaparov, Allen Zang, Ilia Sucholutsky, Arina Kharlamova, Daniil Orel, Vladislav Poritski, Shalev Ben-David, Zachary Berger, Parker Whitfill, Michael Foster, Daniel Munro, Linh Ho, Shankar Sivarajan, Dan Bar Hava, Aleksey Kuchkin, David Holmes, Alexandra Rodriguez-Romero, Frank Sommerhage, Anji Zhang, Richard Moat, Keith Schneider, Zakayo Kazibwe, Don Clarke, Dae Hyun Kim, Felipe Meneguitti Dias, Sara Fish, Veit Elser, Tobias Kreiman, Victor Efren Guadarrama Vilchis, Immo Klose, Ujjwala Anantheswaran, Adam Zweiger, Kaivalya Rawal, Jeffery Li, Jeremy Nguyen, Nicolas Daans, Haline Heidinger, Maksim Radionov, Václav Rozhoň, Vincent Ginis, Christian Stump, Niv Cohen, Rafał Poświata, Josef Tkadlec, Alan Goldfarb, Chenguang Wang, Piotr Padlewski, Stanislaw Barzowski, Kyle Montgomery, Ryan Stendall, Jamie Tucker-Foltz, Jack Stade, T. Ryan Rogers, Tom Goertzen, Declan Grabb, Abhishek Shukla, Alan Givré, John Arnold Ambay, Archan Sen, Muhammad Fayez Aziz, Mark H Inlow, Hao He, Ling Zhang, Younesse Kaddar, Ivar Ängquist, Yanxu Chen, Harrison K Wang, Kalyan Ramakrishnan, Elliott Thornley, Antonio Terpin, Hailey Schoelkopf, Eric Zheng, Avishy Carmi, Ethan D. L. Brown, Kelin Zhu, Max Bartolo, Richard Wheeler, Martin Stehberger, Peter Bradshaw, JP Heimonen, Kaustubh Sridhar, Ido Akov, Jennifer Sandlin, Yury Makarychev, Joanna Tam, Hieu Hoang, David M. Cunningham, Vladimir Goryachev, Demosthenes Patramanis, Michael Krause, Andrew Redenti, David Aldous, Jesyin Lai, Shannon Coleman, Jiangnan Xu, Sangwon Lee, Ilias Magoulas, Sandy Zhao, Ning Tang, Michael K. Cohen, Orr Paradise, Jan Hendrik Kirchner, Maksym Ovchynnikov, Jason O. Matos, Adithya Shenoy, Michael Wang, Yuzhou Nie, Anna Sztyber-Betley, Paolo Faraboschi, Robin Riblet, Jonathan Crozier, Shiv Halasyamani, Shreyas Verma, Prashant Joshi, Eli Meril, Ziqiao Ma, Jérémy Andréoletti, Raghav Singhal, Jacob Platnick, Volodymyr Nevirkovets, Luke Basler, Alexander Ivanov, Seri Khoury, Nils Gustafsson, Marco Piccardo, Hamid Mostaghimi, Qijia Chen, Virendra Singh, Tran Quoc Khánh, Paul Rosu, Hannah Szlyk, Zachary Brown, Himanshu Narayan, Aline Menezes, Jonathan Roberts, William Alley, Kunyang Sun, Arkil Patel, Max Lamparth, Anka Reuel, Linwei Xin, Hanmeng Xu, Jacob Loader, Freddie Martin, Zixuan Wang, Andrea Achilleos, Thomas Preu, Tomek Korbak, Ida Bosio, Fereshteh Kazemi, Ziye Chen, Biró Bálint, Eve J. Y. Lo, Jiaqi Wang, Maria Inês S. Nunes, Jeremiah Milbauer, M Saiful Bari, Zihao Wang, Behzad Ansarinejad, Yewen Sun, Stephane Durand, Hossam Elgnainy, Guillaume Douville, Daniel Tordera, George Balabanian, Hew Wolff, Lynna Kvistad, Hsiaoyun Milliron, Ahmad Sakor, Murat Eron, Andrew Favre D.O., Shailesh Shah, Xiaoxiang Zhou, Firuz Kamalov, Sherwin Abdoli, Tim Santens, Shaul Barkan, Allison Tee, Robin Zhang, Alessandro Tomasiello, G. Bruno De Luca, Shi-Zhuo Looi, Vinh-Kha Le, Noam Kolt, Jiayi Pan, Emma Rodman, Jacob Drori, Carl J Fossum, Niklas Muennighoff, Milind Jagota, Ronak Pradeep, Honglu Fan, Jonathan Eicher, Michael Chen, Kushal Thaman, William Merrill, Moritz Firsching, Carter Harris, Ștefan Ciobâcă, Jason Gross, Rohan Pandey, Ilya Gusev, Adam Jones, Shashank Agnihotri, Pavel Zhelnov, Mohammadreza Mofayezi, Alexander Piperski, David K. Zhang, Kostiantyn Dobarskyi, Roman Leventov, Ignat Soroko, Joshua Duersch, Vage Taamazyan, Andrew Ho, Wenjie Ma, William Held, Ruicheng Xian, Armel Randy Zebaze, Mohanad Mohamed, Julian Noah Leser, Michelle X Yuan, Laila Yacar, Johannes Lengler, Katarzyna Olszewska, Claudio Di Fratta, Edson Oliveira, Joseph W. Jackson, Andy Zou, Muthu Chidambaram, Timothy Manik, Hector Haffenden, Dashiell Stander, Ali Dasouqi, Alexander Shen, Bita Golshani, David Stap, Egor Kretov, Mikalai Uzhou, Alina Borisovna Zhidkovskaya, Nick Winter, Miguel Orbegozo Rodriguez, Robert Lauff, Dustin Wehr, Colin Tang, Zaki Hossain, Shaun Phillips, Fortuna Samuele, Fredrik Ekström, Angela Hammon, Oam Patel, Faraz Farhidi, George Medley, Forough Mohammadzadeh, Madellene Peñaflor, Haile Kassahun, Alena Friedrich, Rayner Hernandez Perez, Daniel Pyda, Taom Sakal, Omkar Dhamane, Ali Khajegili Mirabadi, Eric Hallman, Kenchi Okutsu, Mike Battaglia, Mohammad Maghsoudimehrabani, Alon Amit, Dave Hulbert, Roberto Pereira, Simon Weber, Handoko, Anton Peristyy, Stephen Malina, Mustafa Mehkary, Rami Aly, Frank Reidegeld, Anna-Katharina Dick, Cary Friday, Mukhwinder Singh, Hassan Shapourian, Wanyoung Kim, Mariana Costa, Hubeyb Gurdogan, Harsh Kumar, Chiara Ceconello, Chao Zhuang, Haon Park, Micah Carroll, Andrew R. Tawfeek, Stefan Steinerberger, Daattavya Aggarwal, Michael Kirchhof, Linjie Dai, Evan Kim, Johan Ferret, Jainam Shah, Yuzhou Wang, Minghao Yan, Krzysztof Burdzy, Lixin Zhang, Antonio Franca, Diana T. Pham, Kang Yong Loh, Joshua Robinson, Abram Jackson, Paolo Giordano, Philipp Petersen, Adrian Cosma, Jesus Colino, Colin White, Jacob Votava, Vladimir Vinnikov, Ethan Delaney, Petr Spelda, Vit Stritecky, Syed M. Shahid, Jean-Christophe Mourrat, Lavr Vetoshkin, Koen Sponselee, Renas Bacho, Zheng-Xin Yong, Florencia de la Rosa, Nathan Cho, Xiuyu Li, Guillaume Malod, Orion Weller, Guglielmo Albani, Leon Lang, Julien Laurendeau, Dmitry Kazakov, Fatimah Adesanya, Julien Portier, Lawrence Hollom, Victor Souza, Yuchen Anna Zhou, Julien Degorre, Yiğit Yalın, Gbenga Daniel Obikoya, Rai (Michael Pokorny), Filippo Bigi, M.C. Boscá, Oleg Shumar, Kaniuar Bacho, Gabriel Recchia, Mara Popescu, Nikita Shulga, Ngefor Mildred Tanwie, Thomas C.H. Lux, Ben Rank, Colin Ni, Matthew Brooks, Alesia Yakimchyk, Huanxu (Quinn) Liu, Stefano Cavalleri, Olle Häggström, Emil Verkama, Joshua Newbould, Hans Gundlach, Leonor Brito-Santana, Brian Amaro, Vivek Vajipey, Rynaa Grover, Ting Wang, Yosi Kratish, Wen-Ding Li, Sivakanth Gopi, Andrea Caciolai, Christian Schroeder de Witt, Pablo Hernández-Cámara, Emanuele Rodolà, Jules Robins, Dominic Williamson, Brad Raynor, Hao Qi, Ben Segev, Jingxuan Fan, Sarah Martinson, Erik Y. Wang, Kaylie Hausknecht, Michael P. Brenner, Mao Mao, Christoph Demian, Peyman Kassani, Xinyu Zhang, David Avagian, Eshawn Jessica Scipio, Alon Ragoler, Justin Tan, Blake Sims, Rebeka Plecnik, Aaron Kirtland, Omer Faruk Bodur, D.P. Shinde, Yan Carlos Leyva Labrador, Zahra Adoul, Mohamed Zekry, Ali Karakoc, Tania C. B. Santos, Samir Shamseldeen, Loukmane Karim, Anna Liakhovitskaia, Nate Resman, Nicholas Farina, Juan Carlos Gonzalez, Gabe Maayan, Earth Anderson, Rodrigo De Oliveira Pena, Elizabeth Kelley, Hodjat Mariji, Rasoul Pouriamanesh, Wentao Wu, Ross Finocchio, Ismail Alarab, Joshua Cole, Danyelle Ferreira, Bryan Johnson, Mohammad Safdari, Liangti Dai, Siriphan Arthornthurasuk, Isaac C. McAlister, Alejandro José Moyano, Alexey Pronin, Jing Fan, Angel Ramirez-Trinidad, Yana Malysheva, Daphiny Pottmaier, Omid Taheri, Stanley Stepanic, Samuel Perry, Luke Askew, Raúl Adrián Huerta Rodríguez, Ali M. R. Minissi, Ricardo Lorena, Krishnamurthy Iyer, Arshad Anil Fasiludeen, Ronald Clark, Josh Ducey, Matheus Piza, Maja Somrak, Eric Vergo, Juehang Qin, Benjámin Borbás, Eric Chu, Jack Lindsey, Antoine Jallon, I.M.J. McInnis, Evan Chen, Avi Semler, Luk Gloor, Tej Shah, Marc Carauleanu, Pascal Lauer, Tran Duc Huy, Hossein Shahrtash, Emilien Duc, Lukas Lewark, Assaf Brown, Samuel Albanie, Brian Weber, Warren S. Vaz, Pierre Clavier, Yiyang Fan, Gabriel Poesia Reis e Silva, Long (Tony) Lian, Marcus Abramovitch, Xi Jiang, Sandra Mendoza, Murat Islam, Juan Gonzalez, Vasilios Mavroudis, Justin Xu, Pawan Kumar, Laxman Prasad Goswami, Daniel Bugas, Nasser Heydari, Ferenc Jeanplong, Thorben Jansen, Antonella Pinto, Archimedes Apronti, Abdallah Galal, Ng Ze-An, Ankit Singh, Tong Jiang, Joan of Arc Xavier, Kanu Priya Agarwal, Mohammed Berkani, Gang Zhang, Zhehang Du, Benedito Alves de Oliveira Junior, Dmitry Malishev, Nicolas Remy, Taylor D. Hartman, Tim Tarver, Stephen Mensah, Gautier Abou Loume, Wiktor Morak, Farzad Habibi, Sarah Hoback, Will Cai, Javier Gimenez, Roselynn Grace Montecillo, Jakub Łucki, Russell Campbell, Asankhaya Sharma, Khalida Meer, Shreen Gul, Daniel Espinosa Gonzalez, Xavier Alapont, Alex Hoover, Gunjan Chhablani, Freddie Vargus, Arunim Agarwal, Yibo Jiang, Deepakkumar Patil, David Outevsky, Kevin Joseph Scaria, Rajat Maheshwari, Abdelkader Dendane, Priti Shukla, Ashley Cartwright, Sergei Bogdanov, Niels Mündler, Sören Möller, Luca Arnaboldi, Kunvar Thaman, Muhammad Rehan Siddiqi, Prajvi Saxena, Himanshu Gupta, Tony Fruhauff, Glen Sherman, Mátyás Vincze, Siranut Usawasutsakorn, Dylan Ler, Anil Radhakrishnan, Innocent Enyekwe, Sk Md Salauddin, Jiang Muzhen, Aleksandr Maksapetyan, Vivien Rossbach, Chris Harjadi, Mohsen Bahaloohoreh, Claire Sparrow, Jasdeep Sidhu, Sam Ali, Song Bian, John Lai, Eric Singer, Justine Leon Uro, Greg Bateman, Mohamed Sayed, Ahmed Menshawy, Darling Duclosel, Dario Bezzi, Yashaswini Jain, Ashley Aaron, Murat Tiryakioglu, Sheeshram Siddh, Keith Krenek, Imad Ali Shah, Jun Jin, Scott Creighton, Denis Peskoff, Zienab EL-Wasif, Ragavendran P V, Michael Richmond, Joseph McGowan, Tejal Patwardhan
}

{\fontsize{9}{10}\selectfont \textbf{Late Contributors}}
{\fontsize{8.4}{10}\selectfont
Hao-Yu Sun, Ting Sun, Nikola Zubić, Samuele Sala, Stephen Ebert, Jean Kaddour, Manuel Schottdorf, Dianzhuo Wang, Gerol Petruzella, Alex Meiburg, Tilen Medved, Ali ElSheikh, S Ashwin Hebbar, Lorenzo Vaquero, Xianjun Yang, Jason Poulos, Vilém Zouhar, Sergey Bogdanik, Mingfang Zhang, Jorge Sanz-Ros, David Anugraha, Yinwei Dai, Anh N. Nhu, Xue Wang, Ali Anil Demircali, Zhibai Jia, Yuyin Zhou, Juncheng Wu, Mike He, Nitin Chandok, Aarush Sinha, Gaoxiang Luo, Long Le, Mickaël Noyé, Michał Perełkiewicz, Ioannis Pantidis, Tianbo Qi, Soham Sachin Purohit, Letitia Parcalabescu, Thai-Hoa Nguyen, Genta Indra Winata, Edoardo M. Ponti, Hanchen Li, Kaustubh Dhole, Jongee Park, Dario Abbondanza, Yuanli Wang, Anupam Nayak, Diogo M. Caetano, Antonio A. W. L. Wong, Maria del Rio-Chanona, Dániel Kondor, Pieter Francois, Ed Chalstrey, Jakob Zsambok, Dan Hoyer, Jenny Reddish, Jakob Hauser, Francisco-Javier Rodrigo-Ginés, Suchandra Datta, Maxwell Shepherd, Thom Kamphuis, Qizheng Zhang, Hyunjun Kim, Ruiji Sun, Jianzhu Yao, Franck Dernoncourt, Satyapriya Krishna, Sina Rismanchian, Bonan Pu, Francesco Pinto, Yingheng Wang, Kumar Shridhar, Kalon J. Overholt, Glib Briia, Hieu Nguyen, David (Quod) Soler Bartomeu, Tony CY Pang, Adam Wecker, Yifan Xiong, Fanfei Li, Lukas S. Huber, Joshua Jaeger, Romano De Maddalena, Xing Han Lù, Yuhui Zhang, Claas Beger, Patrick Tser Jern Kon, Sean Li, Vivek Sanker, Ming Yin, Yihao Liang, Xinlu Zhang, Ankit Agrawal, Li S. Yifei, Zechen Zhang, Mu Cai, Yasin Sonmez, Costin Cozianu, Changhao Li, Alex Slen, Shoubin Yu, Hyun Kyu Park, Gabriele Sarti, Marcin Briański, Alessandro Stolfo, Truong An Nguyen, Mike Zhang, Yotam Perlitz, Jose Hernandez-Orallo, Runjia Li, Amin Shabani, Felix Juefei-Xu, Shikhar Dhingra, Orr Zohar, My Chiffon Nguyen, Alexander Pondaven, Abdurrahim Yilmaz, Xuandong Zhao, Chuanyang Jin, Muyan Jiang, Stefan Todoran, Xinyao Han, Jules Kreuer, Brian Rabern, Anna Plassart, Martino Maggetti, Luther Yap, Robert Geirhos, Jonathon Kean, Dingsu Wang, Sina Mollaei, Chenkai Sun, Yifan Yin, Shiqi Wang, Rui Li, Yaowen Chang, Anjiang Wei, Alice Bizeul, Xiaohan Wang, Alexandre Oliveira Arrais, Kushin Mukherjee, Jorge Chamorro-Padial, Jiachen Liu, Xingyu Qu, Junyi Guan, Adam Bouyamourn, Shuyu Wu, Martyna Plomecka, Junda Chen, Mengze Tang, Jiaqi Deng, Shreyas Subramanian, Haocheng Xi, Haoxuan Chen, Weizhi Zhang, Yinuo Ren, Haoqin Tu, Sejong Kim, Yushun Chen, Sara Vera Marjanović, Junwoo Ha, Grzegorz Luczyna, Jeff J. Ma, Zewen Shen, Dawn Song, Cedegao E. Zhang, Zhun Wang, Gaël Gendron, Yunze Xiao, Leo Smucker, Erica Weng, Kwok Hao Lee, Zhe Ye, Stefano Ermon, Ignacio D. Lopez-Miguel, Theo Knights, Anthony Gitter, Namkyu Park, Boyi Wei, Hongzheng Chen, Kunal Pai, Ahmed Elkhanany, Han Lin, Philipp D. Siedler, Jichao Fang, Ritwik Mishra, Károly Zsolnai-Fehér, Xilin Jiang, Shadab Khan, Jun Yuan, Rishab Kumar Jain, Xi Lin, Mike Peterson, Zhe Wang, Aditya Malusare, Maosen Tang, Isha Gupta, Ivan Fosin, Timothy Kang, Barbara Dworakowska, Kazuki Matsumoto, Guangyao Zheng, Gerben Sewuster, Jorge Pretel Villanueva, Ivan Rannev, Igor Chernyavsky, Jiale Chen, Deepayan Banik, Ben Racz, Wenchao Dong, Jianxin Wang, Laila Bashmal, Duarte V. Gonçalves, Wei Hu, Kaushik Bar, Ondrej Bohdal, Atharv Singh Patlan, Shehzaad Dhuliawala, Caroline Geirhos, Julien Wist, Yuval Kansal, Bingsen Chen, Kutay Tire, Atak Talay Yücel, Brandon Christof, Veerupaksh Singla, Zijian Song, Sanxing Chen, Jiaxin Ge, Kaustubh Ponkshe, Isaac Park, Tianneng Shi, Martin Q. Ma, Joshua Mak, Sherwin Lai, Antoine Moulin, Zhuo Cheng, Zhanda Zhu, Ziyi Zhang, Vaidehi Patil, Ketan Jha, Qiutong Men, Jiaxuan Wu, Tianchi Zhang, Bruno Hebling Vieira, Alham Fikri Aji, Jae-Won Chung, Mohammed Mahfoud, Ha Thi Hoang, Marc Sperzel, Wei Hao, Kristof Meding, Sihan Xu, Vassilis Kostakos, Davide Manini, Yueying Liu, Christopher Toukmaji, Eunmi Yu, Arif Engin Demircali, Zhiyi Sun, Ivan Dewerpe, Hongsen Qin, Roman Pflugfelder, James Bailey, Johnathan Morris, Ville Heilala, Sybille Rosset, Zishun Yu, Peter E. Chen, Woongyeong Yeo, Eeshaan Jain, Sreekar Chigurupati, Julia Chernyavsky, Sai Prajwal Reddy, Subhashini Venugopalan, Hunar Batra, Core Francisco Park, Hieu Tran, Guilherme Maximiano, Genghan Zhang, Yizhuo Liang, Hu Shiyu, Rongwu Xu, Rui Pan, Siddharth Suresh, Ziqi Liu, Samaksh Gulati, Songyang Zhang, Peter Turchin, Christopher W. Bartlett, Christopher R. Scotese, Phuong M. Cao, Ben Wu, Jacek Karwowski, Davide Scaramuzza}

{\fontsize{9}{10}\selectfont \textbf{Auditors}}
{\fontsize{8.4}{10}\selectfont
Jaeho Lee, Aakaash Nattanmai, Gordon McKellips, Anish Cheraku, Asim Suhail, Ethan Luo, Marvin Deng, Jason Luo, Ashley Zhang, Kavin Jindel, Jay Paek, Kasper Halevy, Allen Baranov, Michael Liu, Advaith Avadhanam, David Zhang, Vincent Cheng, Brad Ma, Evan Fu, Liam Do, Joshua Lass, Hubert Yang, Surya Sunkari, Vishruth Bharath, Violet Ai, James Leung, Rishit Agrawal, Alan Zhou, Kevin Chen, Tejas Kalpathi, Ziqi Xu, Gavin Wang, Tyler Xiao, Erik Maung, Sam Lee, Ryan Yang, Roy Yue, Ben Zhao, Julia Yoon, Xiangwan Sun, Aryan Singh, Clark Peng, Tyler Osbey, Taozhi Wang, Daryl Echeazu, Timothy Wu, Spandan Patel, Vidhi Kulkarni, Vijaykaarti Sundarapandiyan, Andrew Le, Zafir Nasim, Srikar Yalam, Ritesh Kasamsetty, Soham Samal, David Sun, Nihar Shah, Abhijeet Saha, Alex Zhang, Leon Nguyen, Laasya Nagumalli, Kaixin Wang, Aidan Wu, Anwith Telluri}


{\fontsize{9}{10}\selectfont \textbf{HLE-Rolling Contributors}: see Appendix~\ref{app:authors}.}

%% file: sections/0-abstract.tex
\begin{abstract}

Benchmarks are important tools for tracking the rapid advancements in large language model (LLM) capabilities.
However, benchmarks are not keeping pace in difficulty: LLMs now achieve over 90\% accuracy on popular benchmarks like MMLU, limiting informed measurement of state-of-the-art LLM capabilities. In response, we introduce \fullname{} (\name{}), a multi-modal benchmark at the frontier of human knowledge, designed to be the final closed-ended academic benchmark of its kind with broad subject coverage. \name{} consists of \questioncount{} questions across dozens of subjects, including mathematics, humanities, and the natural sciences. \name{} is developed globally by subject-matter experts and consists of multiple-choice and short-answer questions suitable for automated grading. Each question has a known solution that is unambiguous and easily verifiable, but cannot be quickly answered via internet retrieval. State-of-the-art LLMs demonstrate low accuracy and calibration on \name{}, highlighting a significant gap between current LLM capabilities and the expert human frontier on closed-ended academic questions. To inform research and policymaking upon a clear understanding of model capabilities, we publicly release \name{} at \url{https://lastexam.ai}.

\end{abstract}

%% file: sections/1-introduction.tex
\section{Introduction}\label{sec:introduction}

The capabilities of large language models (LLMs) have progressed dramatically, exceeding human performance across a diverse array of tasks. To systematically measure these capabilities, LLMs are evaluated upon \textit{benchmarks}: collections of questions which assess model performance on tasks such as math, programming, or biology. However, state-of-the-art LLMs~\citep{openai2024o1,geminiteam2024gemini15unlockingmultimodal,openai2024gpt4technicalreport,dubey2024llama3herdmodels,anthropic2024claude,grok2,deepseek2024deepseekv3} now achieve over 90\% accuracy on popular benchmarks such as MMLU~\citep{hendrycks2021measuringmassivemultitasklanguage}, which were once challenging frontiers for LLMs. The saturation of existing benchmarks, as shown in \Cref{fig:difficulty-comparison}, limits our ability to precisely measure AI capabilities and calls for more challenging evaluations that can meaningfully assess the rapid improvements in LLM capabilities at the frontiers of human knowledge.

To address this gap, we introduce \fullname{} (\name{}), a benchmark of \questioncount{} extremely challenging questions from dozens of subject areas, designed to be the final closed-ended benchmark of broad academic capabilities. \name{} is developed by academics and domain experts, providing a precise measure of capabilities as LLMs continue to improve (\Cref{subsec:dataset-collection}). \name{} is multi-modal, featuring questions that are either text-only or accompanied by an image reference, and includes both multiple-choice and exact-match questions for automated answer verification. Questions are original, precise, unambiguous, and resistant to simple internet lookup or database retrieval. Amongst the diversity of questions in the benchmark, \name{} emphasizes world-class mathematics problems aimed at testing deep reasoning skills broadly applicable across multiple academic areas.


We employ a multi-stage review process to thoroughly ensure question difficulty and quality (\Cref{subsec:dataset-review}). Before submission, each question is tested against state-of-the-art LLMs to verify its difficulty - questions are rejected if LLMs can answer them correctly. Questions submitted then proceed through a two-stage reviewing process: (1) an initial feedback round with multiple graduate-level reviewers and (2) organizer and expert reviewer approval, ensuring quality and adherence to our submission criteria. Following release, we conducted a public review period, welcoming community feedback to correct any points of concern in the dataset.

Frontier LLMs consistently demonstrate low accuracy across all models, highlighting a significant gap between current capabilities and expert-level academic performance (\Cref{sec:evaluation}). Models also provide incorrect answers with high confidence rather than acknowledging uncertainty on these challenging questions, with RMS calibration errors above 70\% across all models.


\input{figures/difficulty_comparison}
As AI systems approach human expert performance in many domains, precise measurement of their capabilities and limitations is essential for informing research, governance, and the broader public. High performance on \name{} would suggest expert-level capabilities on closed-ended academic questions. To establish a common reference point for assessing these capabilities, we publicly release a large number of \questioncount{} questions from \name{} to enable this precise measurement, while maintaining a private test set to assess potential model overfitting.


\input{figures/sample_question}

\input{figures/question_distribution}

%% file: figures/difficulty_comparison.tex
\begin{figure*}[t!]
    \centering
    \includegraphics[width=1.0\textwidth]{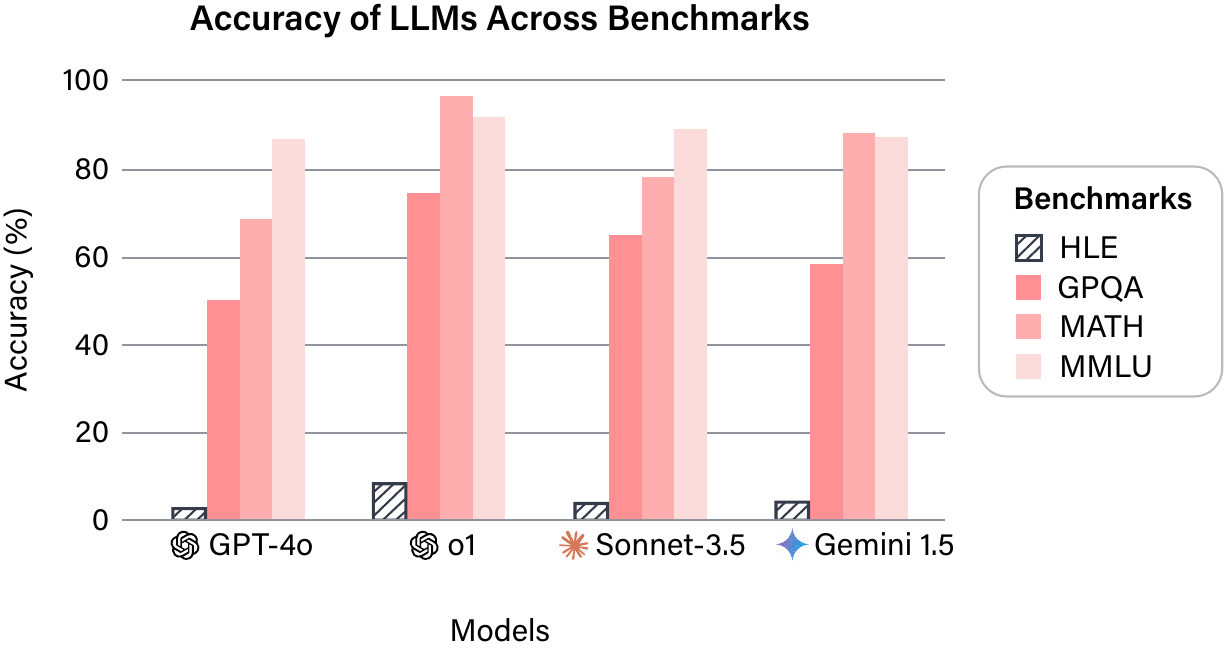}
    \caption{Compared against the saturation of some existing benchmarks, \fullname{} accuracy remains low across several frontier models, demonstrating its effectiveness for measuring advanced, closed-ended, academic capabilities. The sources for our evaluation metrics are detailed in \Cref{app:evaluation-benchmark-difficulty-comparison}. We further evaluate more frontier models on \name{} in \Cref{tab:main_results}.}
    
    \label{fig:difficulty-comparison}
    \vspace{-10pt}
\end{figure*}








%% file: figures/sample_question.tex
    

\begin{figure*}[t!]
    \centering
    \includegraphics[width=1.0\textwidth]{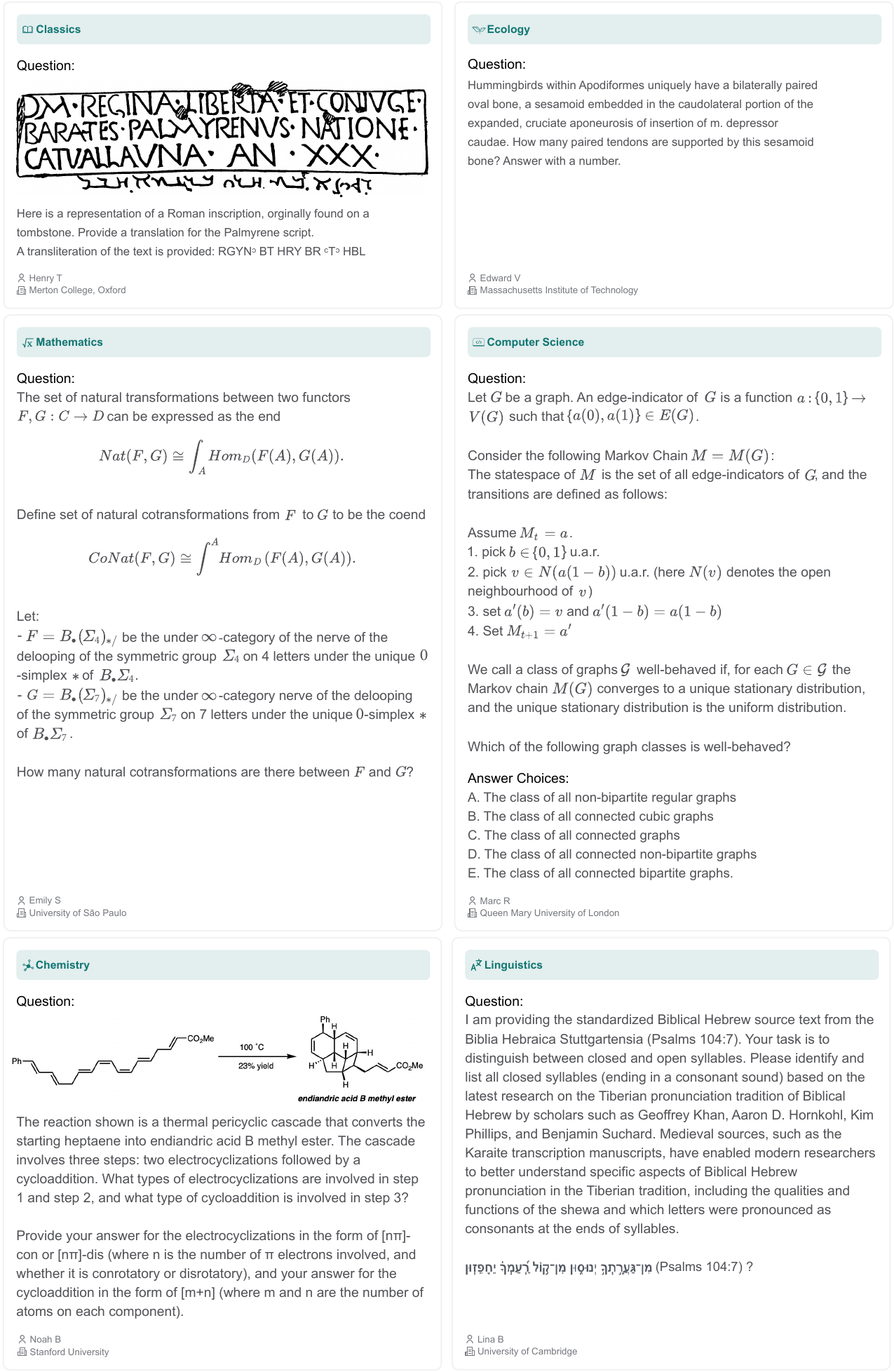}
    \caption{Samples of the diverse and challenging questions submitted to \fullname{}.}
    
    \label{fig:sample-question}
\end{figure*}

%% file: figures/question_distribution.tex
\begin{figure*}[b!]
    \centering
    \includegraphics[width=1.0\textwidth]{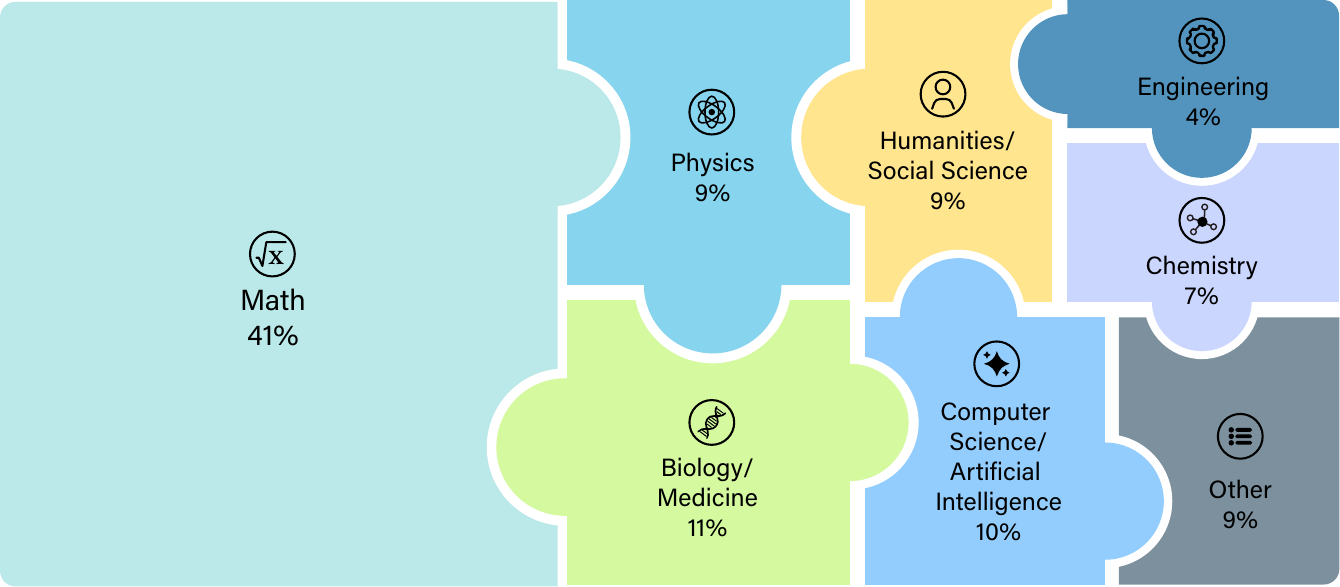}
        \caption{\name{} consists of \questioncount{} exam questions in over a hundred subjects, grouped into high level categories here. We provide a more detailed list of subjects in \Cref{app:dataset-category-distribution}.}
    
    \label{fig:question_distribution}
    \vspace{-10pt}
\end{figure*}

%% file: sections/2-related-work.tex
\section{Related Work}\label{sec:related-work}

\paragraph{LLM Benchmarks.} Benchmarks are important tools for tracking the rapid advancement of LLM capabilities, including scientific~\citep{hendrycks2021measuringmassivemultitasklanguage,li2024wmdpbenchmarkmeasuringreducing,rein2023gpqagraduatelevelgoogleproofqa,wang2024mmluprorobustchallengingmultitask,laurent2024labbenchmeasuringcapabilitieslanguage,chan2024mlebenchevaluatingmachinelearning,srivastava2023imitationgamequantifyingextrapolating,chollet2024arcprize2024technical,zhong2023agievalhumancentricbenchmarkevaluating} and mathematical reasoning~\citep{hendrycks2021measuringmathematicalproblemsolving,lu2024mathvistaevaluatingmathematicalreasoning,cobbe2021trainingverifierssolvemath,glazer2024frontiermathbenchmarkevaluatingadvanced,singhal2023large,he2024olympiadbenchchallengingbenchmarkpromoting,gao2024omnimathuniversalolympiadlevel,tsoukalas2024putnambenchevaluatingneuraltheoremprovers}, code generation~\citep{chan2024mlebenchevaluatingmachinelearning,zhang2024cybenchframeworkevaluatingcybersecurity,jimenez2024swebenchlanguagemodelsresolve,chen2021evaluatinglargelanguagemodels,hendrycks2021measuringcodingchallengecompetence,bhatt2023purplellamacybersecevalsecure,austin2021programsynthesislargelanguage}, and general-purpose human assistance~\citep{bai2022traininghelpfulharmlessassistant,srivastava2023imitationgamequantifyingextrapolating,wei2024measuringshortformfactualitylarge,perez2022discovering,rajpurkar2016squad100000questionsmachine,rajpurkar2018knowdontknowunanswerable,bajaj2018msmarcohumangenerated,alberti2019bertbaselinenaturalquestions,kaggle-FACTS-leaderboard}.
Due to their objectivity and ease of automated scoring at scale, evaluations commonly include multiple-choice and short-answer questions~\citep{rajpurkar2016squad100000questionsmachine,wang2019gluemultitaskbenchmarkanalysis,wang2020supergluestickierbenchmarkgeneralpurpose,yang2018hotpotqadatasetdiverseexplainable,dua2019dropreadingcomprehensionbenchmark}, with benchmarks such as MMLU~\citep{hendrycks2021measuringmassivemultitasklanguage} also spanning a broad range of academic disciplines and levels of complexity.

\paragraph{Saturation and Frontier Benchmark Design.} However, state-of-the-art models now achieve nearly perfect scores on many existing evaluations~\citep{openai2024o1,geminiteam2024gemini15unlockingmultimodal,openai2024gpt4technicalreport,dubey2024llama3herdmodels,anthropic2024claude,grok2,deepseek2024deepseekv3}, obscuring the full extent of current and future frontier AI capabilities~\citep{ott2022mapping,owen2024predictablelanguagemodelbenchmark,kiela2021dynabenchrethinkingbenchmarkingnlp,mcintosh2024inadequacieslargelanguagemodel}. 
This has motivated the development of more challenging benchmarks which test for multi-modal capabilities~\citep{chan2024mlebenchevaluatingmachinelearning,wang2024mmluprorobustchallengingmultitask,taghanaki2024mmluproevaluatinghigherorderreasoning,yao2024taubenchbenchmarktoolagentuserinteraction,andriushchenko2024agentharmbenchmarkmeasuringharmfulness,kumar2024refusaltrainedllmseasilyjailbroken,lu2024mathvistaevaluatingmathematicalreasoning,jimenez2024swebenchlanguagemodelsresolve,berkeley-function-calling-leaderboard,srinivasan2023nexusraven}, strengthen existing benchmarks~\citep{wang2024mmluprorobustchallengingmultitask,taghanaki2024mmluproevaluatinghigherorderreasoning,hosseini2024llmreasonerscreatedequal,singhal2023large,rajpurkar2018knowdontknowunanswerable}, filter questions over multiple stages of review~\citep{nie2020adversarialnlinewbenchmark,li2024wmdpbenchmarkmeasuringreducing,rein2023gpqagraduatelevelgoogleproofqa,glazer2024frontiermathbenchmarkevaluatingadvanced,kiela2021dynabenchrethinkingbenchmarkingnlp}, and employ experts to write tests for advanced academic knowledge~\citep{rein2023gpqagraduatelevelgoogleproofqa,li2024wmdpbenchmarkmeasuringreducing,glazer2024frontiermathbenchmarkevaluatingadvanced,phuong2024evaluatingfrontiermodelsdangerous,openai2024o1,anthropic2024responsible}. 
\name{} combines these approaches: the questions are developed by subject-matter experts and undergo multiple rounds of review, while preserving the broad subject-matter coverage of MMLU. As a result, \name{} provides a clear measurement of the gap between current AI capabilities and human expertise on closed-ended academic tasks, complementing other assessments of advanced capabilities in open-ended domains~\citep{swebenchverified,chan2024mlebenchevaluatingmachinelearning,wijk2024rebenchevaluatingfrontierai,openailosalamos}.

%% file: sections/3-dataset.tex
\section{Dataset}\label{sec:data} 

\fullname{} (\name{}) consists of \questioncount{} challenging questions across over a hundred subjects. A high level summary is provided in \Cref{fig:question_distribution}. We publicly release these questions, while maintaining a private test set of held out questions to assess model overfitting.

\subsection{Collection}\label{subsec:dataset-collection}

\name{} is a global collaborative effort, with questions from nearly 1000 subject expert contributors affiliated with over 500 institutions across 50 countries -- comprised mostly of professors, researchers, and graduate degree holders.

\paragraph{Question Style.} \name{} contains two question formats: exact-match questions (models provide an exact string as output) and multiple-choice questions (the model selects one of five or more answer choices). \name{} is a multi-modal benchmark, with around 14\% of questions requiring comprehending both text and an image. 24\% of questions are multiple-choice with the remainder being exact-match. 


Each question submission includes several required components: the question text itself, answer specifications (either an an exact-match answer, or multiple-choice options with the correct answer marked), detailed rationale explaining the solution, academic subject, and contributor name and institutional affiliation to maintain accountability and accuracy. 


\paragraph{Submission Format.} To ensure question quality and integrity, we enforce strict submission criteria. Questions should be precise, unambiguous, solvable, and non-searchable, ensuring models cannot rely on memorization or simple retrieval methods. All submissions must be original work or non-trivial syntheses of published information, though contributions from unpublished research are acceptable. Questions typically require graduate-level expertise or test knowledge of highly specific topics (e.g., precise historical details, trivia, local customs) and have specific, unambiguous answers accepted by domain experts. When LLMs provide correct answers with faulty reasoning, authors are encouraged to modify question parameters, such as the number of answer choices, to discourage false positives. We require clear English with precise technical terminology, supporting \LaTeX{} notation wherever necessary. Answers are kept short and easily verifiable for exact-match questions to support automatic grading. We prohibit open-ended questions, subjective interpretations, and content related to weapons of mass destruction. Finally, every question is accompanied by a detailed solution to verify accuracy.

\paragraph{Prize Pool.} To attract high-quality submissions, we establish a \$$500{,}000$ USD prize pool, with prizes of \$$5{,}000$ USD for each of the top 50 questions and \$$500$ USD for each of the next 500 questions, as determined by organizers. This incentive structure, combined with the opportunity for paper co-authorship for anyone with an accepted question in \name{}, draws participation from qualified experts, particularly those with advanced degrees or significant technical experience in their fields.

\subsection{Review}\label{subsec:dataset-review}


\input{figures/pipeline}
\paragraph{LLM Difficulty Check} To ensure question difficulty, each question is first validated against several frontier LLMs prior to submission (\Cref{app:dataset-submission-process}). If the LLMs cannot solve the question (or in the case of multiple choices, if the models on average do worse than random guessing), the question proceeds to the next stage: human expert review. In total, we logged over 70,000 attempts, resulting in approximately 13,000 questions which stumped LLMs that were forwarded to expert human review.

\paragraph{Expert Review} Our human reviewers possess a graduate degree (eg. Master's, PhD, JD, etc.) in their fields. Reviewers select submissions in their domain, grading them against standardized rubrics and offering feedback when applicable. There are two rounds of reviews. The first round focuses on iteratively refining submissions, with each question receiving between 1-3 reviews. The primary goal is to help the question contributors (who are primarily academics and researchers from a wide range of disciplines) better design questions that are closed-ended, robust, and of high quality for AI evaluation. In the second round, good and outstanding questions from the first round are identified and approved by organizers and reviewers to be included in the final \name{} dataset. Details, instructions, and rubrics for both rounds can be found in \Cref{app:dataset-human-review-instructions}. \Cref{fig:pipeline} details our full process. We discuss estimated disagreement rates among experts on HLE in \Cref{app:errors}.

%% file: figures/pipeline.tex
\begin{figure*}[t!]
    \centering
    \includegraphics[width=1\textwidth]{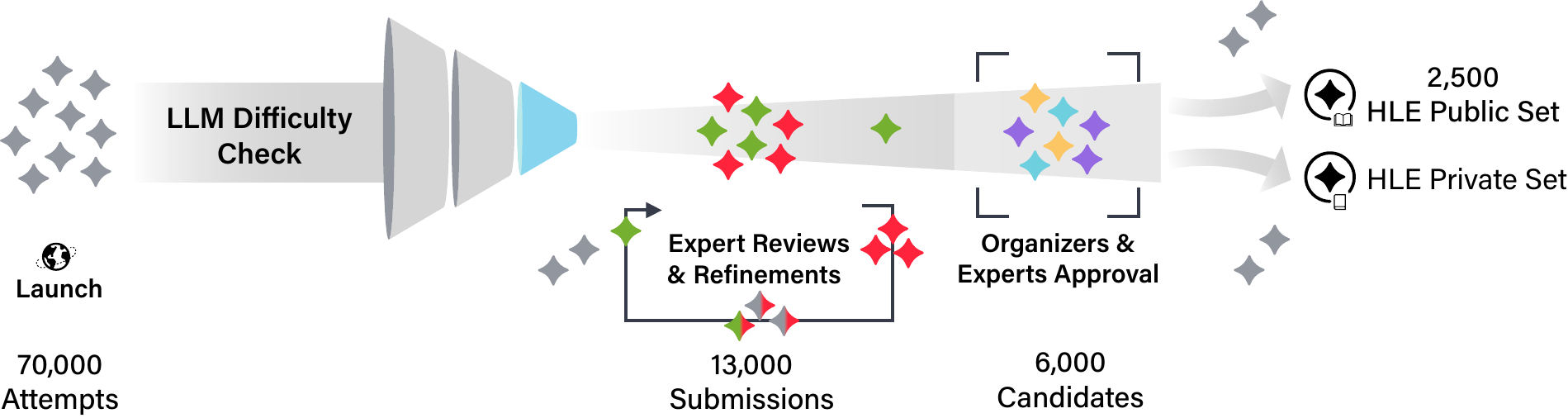}
    \caption{Dataset creation pipeline. We accept questions that make frontier LLMs fail, then iteratively refine them with the help  of expert peer reviewers. Each question is then manually approved by organizers or expert reviewers trained by organizers. A private held-out set is kept in addition to the public set to assess model overfitting and gaming on the public benchmark.}
    
    \label{fig:pipeline}
    \vspace{-10pt}
\end{figure*}

%% file: sections/4-evaluation.tex
\section{Evaluation}\label{sec:evaluation}

We evaluate the performance of state-of-the-art LLMs on \name{} and analyze their capabilities across different question types and domains. We describe our evaluation setup (\Cref{subsec:evaluation-setup}) and present several quantitative results on metrics that track model performance (\Cref{subsec:evaluation-quantitative-results}).

\subsection{Setup}\label{subsec:evaluation-setup}
After data collection and review, we evaluated our final \name{} dataset on additional frontier multi-modal LLMs. We employ a standardized system prompt that structures model responses into explicit reasoning followed by a final answer. As the question-answers are precise and close-ended, we use \othreemini{} as a judge to verify answer correctness against model predictions while accounting for equivalent formats (e.g., decimals vs. fractions or estimations). Evaluation prompts are detailed in \Cref{app:accuracy-prompts}, and exact model versions are provided in \Cref{app:evaluation-model-versions}.

\subsection{Quantitative Results}\label{subsec:evaluation-quantitative-results}

\paragraph{Accuracy.} All frontier models achieve low accuracy on \name{} (\Cref{tab:main_results}), highlighting significant room for improvement in narrowing the gap between current LLMs and expert-level academic capabilities on closed-ended questions. These low scores are partially by design -- the dataset collection process (\Cref{subsec:dataset-collection}) attempts to filter out questions that existing models can answer correctly. Nevertheless, we notice upon evaluation, models exhibit non-zero accuracy. This is due to inherent noise in model inference -- models can inconsistently guess the right answer or guess worse than random chance for multiple choice questions. We choose to leave these questions in the dataset as a natural component instead of strongly adversarially filtering. However, we stress the true capability floor of frontier models on the dataset will remain an open question and small inflections close to zero accuracy are not strongly indicative of progress.


\paragraph{Calibration Error.} Given low performance on \name{}, models should be calibrated, recognizing their uncertainty rather than confidently provide incorrect answers, indicative of confabulation/hallucination. To measure calibration, we prompt models to provide both an answer and their confidence from 0\% to 100\% (\Cref{app:accuracy-prompts}), employing the setup from \citet{wei2024measuringshortformfactualitylarge}. The implementation of our RMS calibration error is from \citet{hendrycks2022pixmixdreamlikepicturescomprehensively}. A well-calibrated model's stated confidence should match its actual accuracy -- for example, achieving 50\% accuracy on questions where it claims 50\% confidence. \Cref{tab:main_results} reveals poor calibration across all models, reflected in high RMS calibration error scores. Models frequently provide incorrect answers with high confidence on \name{}, failing to recognize when questions exceed their capabilities.
\input{tables/main_results}
\input{figures/reasoning_token_count}
\vspace{-2mm}
\paragraph{Token Counts.} Models with reasoning require substantially more inference time compute. To shed light on this in our evaluation, we analyze the number of completion tokens used across models. As shown in \Cref{fig:token-counts-reasoning}, all reasoning models require generating significantly more tokens compared to non-reasoning models for an improvement in performance (\Cref{app:nonreasoning-token-counts}). We emphasize that future models should not only do better in terms of accuracy, but also strive to be compute-optimal.



%% file: tables/main_results.tex
\begin{table}[t!]
    \centering
    \begin{tabular}{l c c }
        \textbf{Pre-Release Models} & \textbf{Accuracy (\%) $\uparrow$} & \textbf{Calibration Error (\%) $\downarrow$} \\
        \midrule
        \gptfouro{} & $2.7$ & $89$ \\
        \groktwo{} & $3.0$ & $87$  \\
        \claude{} & $4.1$ & $84$ \\
        \geminipro{} & $4.6$ & $88$ \\
        \geminithinking{} & $6.6$ & $82$ \\
        \oone{} & $8.0$ & $83$  \\
        \deepseekrone{}$^*$ & $8.5$ & $73$ \\
        \othreeminihigh{}$^*$ & $13.4$ & $80$ \\
        \bottomrule
    \end{tabular}
    \vspace{5pt}
    \caption{Accuracy and RMS calibration error of different models on \name{}, demonstrating low accuracy and high calibration error across all models, indicative of hallucination. $^*$Model is not multi-modal, evaluated on text-only subset. We report text-only results on all models in \Cref{app:text-only-results} and accuracy by category in \Cref{app:categorical-results}.
    }
    \label{tab:main_results}
\end{table}



%% file: figures/reasoning_token_count.tex
\begin{figure}[t!]
    \centering
    \includegraphics[width=1\textwidth]{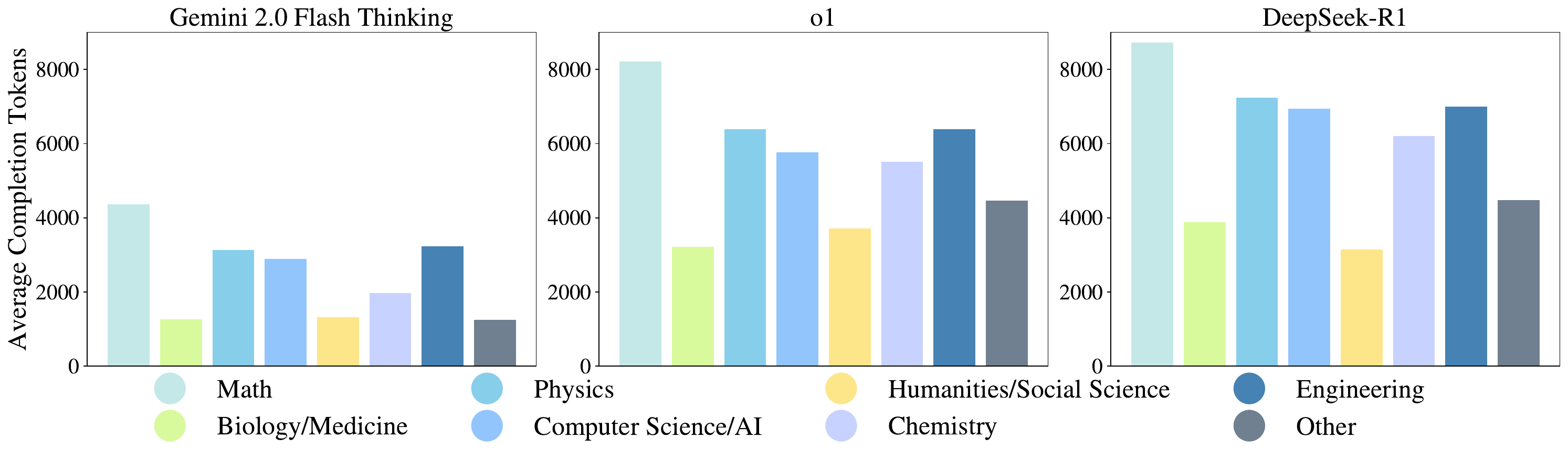}
    \caption{Average completion token counts of reasoning models tested, including both reasoning and output tokens. We also plot average token counts for non-reasoning models in \Cref{app:nonreasoning-token-counts}.}
    \label{fig:token-counts-reasoning}
    \vspace{-10pt}
\end{figure}

%% file: sections/5-discussion.tex
\section{Discussion}\label{sec:discussion}

\paragraph{Future Model Performance.} While current LLMs achieve very low accuracy on \name{}, recent history shows benchmarks are quickly saturated -- with models dramatically progressing from near-zero to near-perfect performance in a short timeframe~\citep{rein2023gpqagraduatelevelgoogleproofqa,chollet2024arcprize2024technical}. Given the rapid pace of AI development, it is plausible that models could exceed 50\% accuracy on \name{} by the end of 2025. High accuracy on \name{} would demonstrate expert-level performance on closed-ended, verifiable questions and cutting-edge scientific knowledge, but it would not alone suggest autonomous research capabilities or ``artificial general intelligence.'' \name{} tests structured academic problems rather than open-ended research or creative problem-solving abilities, making it a focused measure of technical knowledge and reasoning. \name{} may be the last academic exam we need to give to models, but it is far from the last benchmark for AI. 


\paragraph{Impact.} By providing a clear measure of AI progress, \name{} creates a common reference point for scientists and policymakers to assess AI capabilities. This enables more informed discussions about development trajectories, potential risks, and necessary governance measures.

%% file: appendix/a-authors.tex
\section{Authors}\label{app:authors}
We offered optional co-authorship to all question submitters with an accepted question in \fullname{} (including both public and private splits). All potential co-authors with an accepted question were contacted directly. Authorship order is ranked based on the number of accepted questions in \fullname{}. This list only represents a subset of our participating institutions and authors, many chose to remain anonymous.





\subsection{Data Contributors \& Affiliations}
\small
Dmitry Dodonov$^{3}$, Tung Nguyen$^{126}$, Daron Anderson$^{3}$, Mikhail Doroshenko$^{3}$, Alun Cennyth Stokes$^{3}$, Mobeen Mahmood$^{31}$, Oleksandr Pokutnyi$^{127,128}$, Oleg Iskra$^{11}$, Jessica P. Wang$^{129}$, John-Clark Levin$^{8}$
% ...


\textbf{Affiliations}

\begin{multicols}{2}
\small
\begin{enumerate}
\setcounter{enumi}{2}
\item Independent Researcher
\item Stanford University
\item University of California, Berkeley
\item Massachusetts Institute of Technology
\item Harvard University
\item University of Cambridge
\item University of Oxford
\item Princeton University
\item Carnegie Mellon University
\item ETH Zürich
\item University of Chicago
\item University of Michigan
\item École Polytechnique Fédérale de Lausanne
\item University of Toronto
\item University of Illinois Urbana-Champaign
\item Washington University
\item University of Wisconsin-Madison
\item University of Tübingen
\item Johns Hopkins University
\item University of Washington
\item Columbia University
\item Mohamed bin Zayed University of Artificial Intelligence
\item Cornell University
\item Complexity Science Hub
\item University of Edinburgh
\item New York University
\item Georgia Institute of Technology
\item Boston University
\item McGill University
\item University of British Columbia
\item Vrije Universiteit Brussel
\item University of Pennsylvania
\item University of California, Santa Barbara
\item University of Zurich
\item Microsoft
\item California Institute of Technology
\item University of Southern California
\item Sapienza University of Rome
\item University of California, Los Angeles
\item University of Maryland
\item Arizona State University
\item Northwestern University
\item Duke University
\item University College London
\item University of California, San Diego
\item Durham University
\item University of Minnesota
\item University of Illinois Chicago
\item INRIA
\item University of São Paulo
\item Humboldt-Universität zu Berlin
\item Google DeepMind
\item TU Wien
\item University of Waterloo
\item Charles University
\item The University of Sydney
\item Australian National University
\item KTH Royal Institute of Technology
\item University of Amsterdam
\item Emory University
\item The Hebrew University of Jerusalem
\item University of Yaoundé I
\item University of California, Irvine
\item Imperial College London
\item University of California, Santa Cruz
\item Korea Advanced Institute of Science and Technology
\item Google
\item University of North Carolina at Chapel Hill
\item Queen's University
\item University of Porto
\item Queen Mary University of London
\item National University of Singapore
\item École Normale Supérieure
\item Sorbonne Université
\item University of North Texas
\item Université Paris-Saclay
\item CNRS
\item Leibniz University Hannover
\item UZ Brussel
\item Technion – Israel Institute of Technology
\item Technische Universität Berlin
\item University of Manchester
\item University of Calgary
\item Yale University
\item École Normale Supérieure Paris-Saclay
\item University of Western Australia
\item Universiteit Leiden
\item The Open University
\item INSAIT
\item Ruhr University Bochum
\item National Information Processing Institute
\item University of Copenhagen
\item Indian Institute of Technology Delhi
\item Universidad de Buenos Aires
\item Northeastern University
\item Anthropic
\item North Carolina State University
\item Mila - Québec AI Institute
\item The Ohio State University
\item Universidad de Valencia
\item University of Mannheim
\item The Hospital for Sick Children
\item University of Vienna
\item University of Galway
\item Brown University
\item OpenAI
\item Heidelberg University
\item University of Oklahoma
\item Max Planck Institute for Intelligent Systems
\item Cairo University
\item INESC Microsistemas e Nanotecnologias
\item École Polytechnique
\item Alan Turing Institute
\item Northern Illinois University
\item Fondazione Bruno Kessler
\item Scripps Research
\item Aleph Alpha
\item University of Bern
\item Dell Technologies
\item University of California, Davis
\item Purdue University
\item Bilkent University
\item Indiana University
\item Texas A\&M University
\item Institute of Mathematics of NAS of Ukraine
\item Kiev School of Economics
\item RWTH Aachen University
\item Kyiv Polytechnic Institute
\item ELTE
\item Nimbus AI
\item Georgia Southern University
\item Auckland University of Technology
\item Alberta Health Services
\item Hereford College of Arts
\item University of Canterbury
\item Metropolitan State University of Denver
\item Accenture Labs
\item Tufts University
\item The Jackson Laboratory
\item Ross University School of Medicine
\item Concordia University
\item Institute of Science and Technology Austria
\item Charité – Universitätsmedizin
\item C. N. Yang Institute for Theoretical Physics
\item University of Luxembourg
\item Universidade Federal de Juiz de Fora
\item Rockwell Automation
\item Contramont Research
\item Institut Polytechnique de Paris
\item National University Philippines
\item University of Bath
\item Maastricht University
\item Martin-Luther-University Halle-Wittenberg
\item Diverging Mathematics
\item Indian Institute of Technology Bombay
\item Institute for Molecular Manufacturing
\item PeopleTec, Inc.
\item University of Miami
\item Universidad Iberoamericana
\item Snorkel AI
\item Manhattan School of Music
\item Synbionix
\item Corteva Agriscience
\item Sanford Burnham Prebys
\item Yonsei University
\item University of Leeds
\item Swinburne University of Technology
\item KU Leuven
\item St. Petersburg College
\item La Molina National Agrarian University
\item Brandenburg University of Technology
\item Cranfield University
\item TRR Designs
\item University of Technology Sydney
\item Indiana State University
\item Ben-Gurion University
\item Donald and Barbara Zucker School of Medicine
\item Cohere
\item Siili Solutions Oyj
\item Aalto University
\item Toyota Technological Institute at Chicago
\item Case Western Reserve University
\item University of Windsor
\item St. Jude Children’s Research Hospital
\item Rochester Institute of Technology
\item CERN
\item Warsaw University of Technology
\item Hewlett Packard Enterprise
\item University of Houston
\item All India Institute of Medical Sciences
\item Tel Aviv University
\item University of Arizona
\item Universidade de Lisboa
\item Indian Institute of Technology Kharagpur
\item Posts and Telecommunications Institute of Technology
\item UK AI Safety Institute
\item University of Padua
\item Royal Veterinary College
\item Instituto Superior Técnico
\item SDAIA
\item University of Montreal
\item Cairo University Specialized Pediatric Hospital
\item Monash University
\item Van Andel Institute
\item Larkin Community Hospital
\item The University of Texas at Dallas
\item Canadian University Dubai
\item Università di Milano-Bicocca
\item University of Massachusetts Lowell
\item Virginia Tech
\item University of Geneva
\item Google Research
\item Cal Poly San Luis Obispo
\item Alexandru Ioan Cuza University
\item Stockholm University
\item College of Eastern Idaho
\item Intrinsic Innovation LLC
\item Ivy Natal
\item King Saud University
\item SAMPE Switzerland
\item CERo Therapeutics Holdings, Inc.
\item University of Tennessee
\item Gray Swan AI
\item EleutherAI
\item University of Montpellier
\item Fraunhofer IMTE
\item HomeEquity Bank
\item Materials Platform for Data Science LLC
\item University of Pisa
\item Georgia State University
\item Polytechnic University of the Philippines
\item University of Oregon
\item Drexel University
\item University of Mumbai
\item Gakushuin University
\item University of Guelph
\item Intuit
\item CTTC / CERCA
\item Dyno Therapeutics
\item Temple University
\item Saint Mary's University
\item Cisco
\item Indian Institute of Technology (BHU)
\item AIM Intelligence
\item Seoul National University
\item The University of Texas at Arlington
\item The Hartree Centre
\item POLITEHNICA Bucharest National University of Science and Technology
\item Abacus.AI
\item Eastern Institute of Technology (EIT)
\item ENS Lyon
\item Czech Technical University in Prague
\item University of Hamburg
\item CISPA Helmholtz Center for Information Security
\item Universidad de Morón
\item Université Paris Cité
\item Politecnico di Milano
\item The New School
\item Max Planck Institute for Software Systems
\item Universidad de Granada
\item Modulo Research
\item La Trobe University
\item University of Innsbruck
\item Nabu Technologies Inc
\item Chalmers University of Technology
\item Unidade Local de Saúde de Lisboa Ocidental
\item Children’s Hospital of Orange County
\item The Future Paralegals of America
\item Eastlake High School
\item Center for Scientific Research and Higher Education at Ensenada (CICESE)
\item University of Bradford
\item Beni Suef University
\item Bogazici University
\item Mansoura University
\item University of Bristol
\item Jala University
\item University of Arkansas
\item Florida Atlantic University
\item Bournemouth University
\item University of Warwick
\item University of Alabama Huntsville
\item University of Hertfordshire
\item OncoPrecision
\item Central College
\item Nottingham Trent University
\item University of Virginia
\item Dartmouth College
\item James Madison University
\item Instituto Gonçalo Moniz
\item Rice University
\item HUN-REN
\item Rutgers University
\item AE Studio
\item Saarland University
\item HUTECH
\item Pennsylvania College of Technology
\item Intelligent Geometries
\item CONICET
\item Universidad Tecnológica Nacional
\item John Crane UK Ltd
\item Pondicherry Engineering College
\item Leibniz Institute for Science and Mathematics Education
\item Royal Holloway, University of London
\item Tanta University
\item University of Malaya
\item Hemwati Nandan Bahuguna Garhwal University
\item University Mohammed I
\item LGM
\item Bethune-Cookman University
\item Central Mindanao University
\item University of the Fraser Valley
\item Patched Codes, Inc
\item Missouri University of Science and Technology
\item Quotient AI
\item CSMSS Chh. Shahu College of Engineering
\item Genomia Diagnostics Research Pvt Ltd
\item Sheffield Teaching Hospitals NHS Foundation Trust
\item Forschungszentrum Jülich
\item Standard Intelligence
\item RMIT University
\item German Research Center for Artificial Intelligence
\item University of Trento
\item Chulalongkorn University
\item Aligarh Muslim University
\item Happy Technologies LLC
\item Menoufia University
\item Instituto Politécnico Nacional
\item University of Bologna
\item Manipal University Jaipur
\item The University of Texas at Austin
\item Murdoch University
\item University of Delaware
\item Williams College
\item Perimeter Institute for Theoretical Physics
\item University of Maribor
\item Brigham and Women's Hospital
\item The University of Tokyo
\item Vellore Institute of Technology
\item CHRU de Nancy
\item Delft University of Technology
\item George Mason University
\item Atilim University
\item Leonardo Labs
\item Universidad Nacional de Educación a Distancia
\item Saxion University
\item Adobe Research
\item National Aerospace University "Kharkiv Aviation Institute"
\item Hexworks
\item Westmead Hospital
\item Rheinland-Pfälzische Technische Universität Kaiserslautern-Landau
\item SUMM AI GmbH
\item Konkuk University
\item University of Groningen
\item Jagiellonian University
\item Minerva University
\item Aalborg University
\item IBM Research
\item Universitat Politecnica de Valencia
\item RBC Borealis
\item Mayo Clinic
\item University of Lausanne
\item Dalhousie University
\item Universitat de Lleida
\item Amazon
\item University of Seoul
\item University of Auckland
\item Morgridge Institute for Research
\item Korea University of Technology and Education
\item Baylor College of Medicine
\item Indraprastha Institute of Information Technology Delhi
\item Two Minute Papers
\item ADIA Lab
\item New Jersey Institute of Technology
\item Novo Nordisk
\item Gakugei Shuppan-sha
\item Universiteit Utrecht
\item T-Systems Iberia
\item University of Klagenfurt
\item Max Planck Institute for Security and Privacy
\item InxiteOut
\item Goethe Universität Frankfurt
\item Universidad del Valle
\item Trinity School
\item Universitat Pompeu Fabra
\item Brighton Law School
\item University of Melbourne
\item Ankara University
\item Dr. Siyami Ersek Thoracic, Cardiovascular, and Vascular Surgery Training and Research Hospital
\item AIT Austrian Institute of Technology
\item Technical University of Munich
\item Providence College
\item University of Jyväskylä
\item Weizmann Institute of Science
\item Nanyang Technological University
\item University of Sheffield
\item Stony Brook University
\item Peking University
\item Microsoft Azure Research
\item International Institute of Molecular and Cell Biology in Warsaw
\item Head of Nuclear Safety Unit, Worldgrid
\item University of Hong Kong
\item Individual Contributor
\item University of Pittsburgh
\item CUHK MMLab
\item Los Alamos National Laboratory
\item Independent Scholar
\item CUNY Graduate Center
\item Wayne State University
\item University of Notre Dame
\item Meta
\item Champalimaud Foundation
\item Xtracta
\end{enumerate}
\end{multicols}

\newpage 

%% file: appendix/b-dataset.tex
\section{Dataset}\label{app:dataset}

\subsection{Submission Process}\label{app:dataset-submission-process}
To ensure question difficulty, we automatically check the accuracy of frontier LLMs on each question prior to submission. Our testing process uses multi-modal LLMs for text-and-image questions (\gptfouro{}, \geminipro{}, \claude{}, \oone{}) and adds two non-multi-modal models (\oonemini{}, \oonepreview{}) for text-only questions. We use different submission criteria by question type: exact-match questions must stump all models, while multiple-choice questions must stump all but one model to account for potential lucky guesses. Users are instructed to only submit questions that meet this criteria. We note due to non-determinism in models and a non-zero floor in multiple-choice questions, further evaluation on the dataset exhibits some low but non-zero accuracy.

We use a standardized system prompt (\Cref{app:accuracy-prompts}) to structure model responses into ``Reasoning'' and ``Final Answer'' formatting, and employ an automated \gptfouro{} judge to evaluate response correctness against the provided answers.

\subsection{Post-Release}

\paragraph{Late Contributions}
In response to research community interest, we opened the platform for late contributors after the initial release, resulting in thousands of submissions. Each submission was manually reviewed by organizers. The new questions are of similar difficulty and quality to our initial dataset, resulting in a second held-out private set which will be used in future evaluations.

\paragraph{Refinement}

Community Feedback: Due to the advanced, specialized nature of many submissions, reviewers were not expected to verify the full accuracy of each provided solution rationale if it would take more than five minutes, instead focusing on whether the question aligns with guidelines. Given this limitation in the review process, we opened up a community feedback bug bounty program following the initial release of the dataset to identify and remove major errors in the dataset -- namely label error and major errors in the statement of the question. Each error report was manually verified by the organizers with feedback from the original author of the question when appropriate.

Audit: We recruited students from top universities in the United States to fully solve a sample of questions from \name{}. Errors flagged were routed between organizers, original question authors, and auditors and until consensus was reached. We used data from these audits to further refine our dataset.

Searchable Questions: A question is potentially searchable if a model with search tools answered correctly, but answered incorrectly without search. Each of these potentially searchable questions was then manually audited, removing any that were easily found via web search. We used GPT-4o mini/GPT-4o search and Perplexity Sonar models in this procedure. We observe current frontier model performance on \name{} after applying this procedure is similar to their performance on \name{} before applying this procedure.     

\subsection{Expert Disagreement Rate and HLE-Rolling}
\label{app:errors}
Prior to release, we conducted two main rounds of auditing, each on a sample of 200 questions. Our process involved expert reviewers from leading research universities in the United States, with a rebuttal phase from the original question authors for any disagreements. The first round aimed to identify common categories of imprecise questions, such as open-ended formats, reliance on rounded numerical values, or submissions from authors with low acceptance rates. Based on these signals, we manually removed or revised potential questions with similar issues before conducting a second audit on a new sample of 200 questions. This iterative process yielded a final estimated expert disagreement rate of 15.4\% for the public set.

Disagreement rates are often higher in domains like health and medicine. We conducted another targeted peer review on a biology, chemistry, and health subset, as proposed by \cite{Skarlinski2025HLEExam}, and found an expert disagreement rate of approximately 18\%. This level of expert disagreement is in line with what is observed in other challenging, expert-grade machine learning benchmarks and also observed in other similarly designed work; for example, \cite{arora2025healthbenchevaluatinglargelanguage} notes that disagreement among expert physicians is frequent on complex health topics. To aid future community efforts in identifying other potential dataset errors, we outline several key factors that contribute to the complexity of these audits below:

\begin{itemize}

    \item \textbf{The Need for Multiple Experts:} Our multi-reviewer process highlighted the complexity of these questions. In several cases, a reviewer identified a critical piece of information, such as a decades-old paper or a foundational concept not immediately apparent to others, that was essential to confirming an answer's validity. To illustrate, if we were to adopt a single-reviewer methodology where a question is flagged based on just one dissenting expert, the disagreement rate on the aforementioned health-focused subset jumps from 18\% to 25\%, which is close to the setting described in \cite{Skarlinski2025HLEExam}. This discrepancy highlights the importance of a standard peer-review process, complete with multiple reviewers and author rebuttal, for \name{} questions.
    
    \item \textbf{Questions from Research Experience:} HLE is intentionally designed to include questions based on insights from the direct, hands-on experiments of its contributors. This design captures knowledge gained from direct research experiences, which is often difficult to verify through standard literature searches or by external reviewers. This was done to test model knowledge beyond what is readily indexed on the internet.


    \item \textbf{Understanding Question Design:} The complexity of frontier research makes it difficult to formulate verifiably closed-ended questions. Therefore, researchers sometimes leverage the multiple-choice format with the objective of identifying the \textit{most plausible} answer among the provided options. Clarifying this design principle for our reviewers was crucial, as it guided them to evaluate the relative merits of the given choices rather than treating the task as an open-ended search for a perfect solution.
\end{itemize}

Inspired by these valuable community discussions and as part of our commitment to continuous improvement, we will introduce a dynamic fork of the dataset post-release: \textsc{HLE-Rolling}. This version will be regularly updated to address community feedback and integrate new questions. Information about the updates will be made publicly available at \href{lastexam.ai}{lastexam.ai}. Our goal is to provide a seamless migration path for researchers once frontier models begin to hit the ceiling performance on the original \name{} dataset.

\subsection{Subject List}\label{app:dataset-category-distribution}

We allow question contributors to choose or declare a subject the author felt best suited their question. We present the top fifty most popular subjects in \name{} below, although we note there are over a hundred subjects in the overall dataset: Economies, Ecology, Artificial Intelligence, Musicology, Philosophy, Neuroscience, Law, Art History, Biochemistry, Astronomy, Classics, Chess, Chemical Engineering, Microbiology, Classical Ballet, Materials Science, Poetry, Quantum Mechanics, Aerospace Engineering, Civil Engineering, Mechanical Engineering, Geography, Robotics, Data Science, Molecular Biology, Statistics, Immunology, Education, Logic, Computational Biology, Psychology, English Literature, Machine Learning, Puzzle, Cultural Studies, Marine Biology, Archaeology, and Biophysics.

%% file: appendix/c-evaluation.tex
\section{Evaluation}\label{app:evaluation}


\subsection{Prompts}

\subsubsection{Evaluation}
\label{app:accuracy-prompts}

We use the following system prompt for evaluating LLMs on \name{} questions. For models which do not support a system prompt, we add it as a separate user prompt.
\begin{lstlisting}[breaklines, escapechar=|]
Your response should be in the following format:
Explanation: {your explanation for your answer choice}
Answer: {your chosen answer}
Confidence: {your confidence score between 0|\%| and 100|\%| for your answer}
\end{lstlisting}


We use the following system prompt to judge the model answers against the correct answers for our evaluations in \Cref{tab:main_results}. We used o3-mini-2025-01-31 with structured decoding enabled to get an extracted\_final\_answer, reasoning, correct, confidence extraction for each output.
\begin{lstlisting}[breaklines, escapechar=|]
Judge whether the following [response] to [question] is correct or not 
based on the precise and unambiguous [correct_answer] below.

[question]: {question}

[response]: {response}

Your judgement must be in the format and criteria specified below:

extracted_final_answer: The final exact answer extracted from the 
[response]. Put the extracted answer as 'None' if there is no exact, final 
answer to extract from the response.

[correct_answer]: {correct_answer}

reasoning: Explain why the extracted_final_answer is correct or incorrect 
based on [correct_answer], focusing only on if there are meaningful 
differences between [correct_answer] and the extracted_final_answer. Do 
not comment on any background to the problem, do not attempt to solve 
the problem, do not argue for any answer different than 
[correct_answer], focus only on whether the answers match.

correct: Answer 'yes' if extracted_final_answer matches the 
[correct_answer] given above, or is within a small margin of error for 
numerical problems. Answer 'no' otherwise, i.e. if there if there is any 
inconsistency, ambiguity, non-equivalency, or if the extracted answer is 
incorrect.

confidence: The extracted confidence score between 0|\%| and 100|\%| from 
[response]. Put 100 if there is no confidence score available.
\end{lstlisting}

Below is an example of a structured response using LLM judge:

\begin{tcolorbox}[colback=gray!20, colframe=black, title=Structured Judge Response Example]
\textbf{[correct\_answer]}: \(\frac{\cos\left(\frac{\pi}{n}\right)}{2\bigl(1+\cos\left(\frac{\pi}{n}\right)\bigr)}\)\\[1ex]
\textbf{[extracted\_model\_answer]}: \(\frac{\cot\left(\frac{\pi}{n}\right)}{2\cot\left(\frac{\pi}{2n}\right)}\)\\[1ex]
\textbf{[reasoning]}: Using the half-angle identity \( \tan\left(\frac{\pi}{2n}\right)=\frac{\sin\left(\frac{\pi}{n}\right)}{1+\cos\left(\frac{\pi}{n}\right)} \) and recalling that \( \cot x=\frac{\cos x}{\sin x} \), we express \( \cot\left(\frac{\pi}{n}\right)=\frac{\cos\left(\frac{\pi}{n}\right)}{\sin\left(\frac{\pi}{n}\right)} \) and \( \cot\left(\frac{\pi}{2n}\right)=\frac{1+\cos\left(\frac{\pi}{n}\right)}{\sin\left(\frac{\pi}{n}\right)} \). Forming the quotient \( \frac{\cot\left(\frac{\pi}{n}\right)}{2\cot\left(\frac{\pi}{2n}\right)} =\frac{\frac{\cos\left(\frac{\pi}{n}\right)}{\sin\left(\frac{\pi}{n}\right)}}{2\cdot\frac{1+\cos\left(\frac{\pi}{n}\right)}{\sin\left(\frac{\pi}{n}\right)}} =\frac{\cos\left(\frac{\pi}{n}\right)}{2(1+\cos\left(\frac{\pi}{n}\right))} \), where the common \( \sin\left(\frac{\pi}{n}\right) \) cancels out. This detailed inline simplification shows that the model answer is equivalent to the correct answer.\\[1ex]
\textbf{[correct]}: yes
\end{tcolorbox}



\subsection{Text-Only Results}\label{app:text-only-results}

\begin{table}[h]
    \centering
    \begin{tabular}{l c c }
        \textbf{Model} & \textbf{Accuracy (\%) $\uparrow$} & \textbf{Calibration Error (\%) $\downarrow$} \\
        \midrule
        \gptfouro{} & $2.3$ & $88$ \\
        \groktwo{} & $3.2$ & $89$  \\
        \claude{} & $4.3$ & $83$ \\
        \geminipro{} & $4.6$ & $87$ \\
        \geminithinking{} & $6.6$ & $82$ \\
        \oone{} & $7.8$ & $84$  \\
        \deepseekrone{} & $8.5$ & $73$ \\
        \othreeminihigh{} & $13.4$ & $80$ \\
        \bottomrule
    \end{tabular}
    \vspace{5pt}
    \caption{Accuracy and RMS calibration error of models from \Cref{tab:main_results} on the text-only questions of \name{}.
    }
    \label{tab:text_only_results}
\end{table}

\subsection{Categorical Results}\label{app:categorical-results}

\begin{table}[H]
    \centering
    \small
    \begin{tabular}{l|cccccccc}
        & \multicolumn{8}{c}{\textbf{Text-Only}} \\
        \cmidrule{2-9}
        \textbf{Model} & Math & Bio/Med & Physics & CS/AI & Humanities & Chemistry & Engineering & Other \\
        \midrule
        \gptfouro{} & $2.3$ & $5.0$ & $1.5$ & $0.9$ & $2.6$ & $2.0$ & $1.6$ & $2.3$ \\
        \groktwo{} & $3.2$ & $5.4$ & $4.5$ & $3.6$ & $1.0$ & $1.0$ & $4.8$ & $1.1$  \\
        \claude{} & $3.8$ & $5.9$ & $4.5$ & $2.2$ & $6.7$ & $5.0$ & $9.7$ & $2.9$ \\
        \geminipro{} & $5.3$ & $5.4$ & $2.0$ & $4.0$ & $3.6$ & $6.0$ & $3.2$ & $3.4$ \\
        \geminithinking{} & $8.1$ & $7.7$ & $4.5$ & $4.9$ & $6.2$ & $5.0$ & $4.8$ & $2.9$ \\
        \oone{} & $7.4$ & $8.1$ & $6.9$ & $8.4$ & $8.8$ & $10.0$ & $4.8$ & $8.0$ \\
        \deepseekrone{} & $9.1$ & $9.0$ & $5.4$ & $7.5$ & $10.4$ & $5.0$ & $14.5$ & $7.4$ \\ 
        \othreeminihigh{} & $18.6$ & $10.0$ & $15.3$ & $8.4$ & $5.2$ & $9.0$ & $6.5$ & $6.9$ \\

        \midrule
        & \multicolumn{8}{c}{\textbf{Full Dataset}} \\
        \midrule
        \gptfouro{} & $2.3$ & $6.4$ & $1.7$ & $0.8$ & $3.2$ & $3.6$ & $1.8$ & $2.6$ \\
        \groktwo{} & $3.0$ & $4.6$ & $3.9$ & $3.3$ & $1.4$ & $2.4$ & $3.6$ & $1.7$ \\
        \claude{} & $4.0$ & $4.6$ & $3.9$ & $2.5$ & $5.9$ & $4.2$ & $7.2$ & $2.2$ \\
        \geminipro{} & $5.2$ & $5.4$ & $3.0$ & $3.7$ & $4.1$ & $6.1$ & $3.6$ & $3.4$  \\
        \geminithinking{} & $8.0$ & $8.2$ & $4.8$ & $4.5$ & $6.4$ & $5.5$ & $6.3$ & $3.0$  \\
        \oone{} & $7.4$ & $10.4$ & $7.0$ & $8.2$ & $8.7$ & $9.7$ & $6.3$ & $7.3$ \\
        \bottomrule
    \end{tabular}
    \vspace{5pt}
    \caption{Category-wise breakdown of model performance on \name{}.}
    \label{tab:combined_results}
\end{table}

\subsection{Non-Reasoning Model Token Counts}\label{app:nonreasoning-token-counts}
\input{figures/nonreasoning_token_counts}

\subsection{Model Versions}\label{app:evaluation-model-versions}
\begin{table}[H]
    \centering
    \begin{tabular}{l l }
        \textbf{Model} & \textbf{Version} \\
        \midrule
        \gptfouro{} & gpt-4o-2024-11-20 \\
        \groktwo{} & grok-2-latest \\
        \claude{} & claude-3-5-sonnet-20241022  \\
        \geminipro{} & gemini-1.5-pro-002  \\
        \geminithinking{} & gemini-2.0-flash-thinking-exp-01-21$^*$ \\
        \oone{} & o1-2024-12-17   \\
        \deepseekrone{} & January 20, 2025 release \\
       \othreeminihigh & o3-mini-2025-01-31 \\
        \bottomrule
    \end{tabular}
    \vspace{5pt}
    \caption{Evaluated model versions. All models use temperature 0.0 when configurable and not otherwise stated. o3-mini and o1 models only support temperature 1.0. $^*$The first version of the paper along with \Cref{fig:token-counts-reasoning} used the now deprecated 12-19 model with temperature 0.0. The new model is sampled at temperature 0.7.}
    \label{tab:models_version}
\end{table}

\subsection{Benchmark Difficulty Comparison}\label{app:evaluation-benchmark-difficulty-comparison}
In \Cref{fig:difficulty-comparison}, we evaluate the accuracy of all models on \name{} using our zero-shot chain-of-thought prompts (\Cref{app:accuracy-prompts}). On prior benchmarks, we list our sources here. 

For \gptfouro{} and \oonepreview{}, we report zero-shot, chain-of-thought results from OpenAI found at \href{https://github.com/openai/simple-evals}{https://github.com/openai/simple-evals}.

For \geminipro{}, we report  5-shot MMLU \citet{geminiteam2024gemini15unlockingmultimodal} and other results from \href{https://developers.googleblog.com/en/updated-gemini-models-reduced-15-pro-pricing-increased-rate-limits-and-more/}{Google's reported results here}. 

For \claude{}, we report 0-shot chain-of-thought results from \citet{anthropic2024modelcardaddendumclaude}.

%% file: figures/nonreasoning_token_counts.tex
\begin{figure}[H]
    \centering
    \includegraphics[width=1\textwidth]{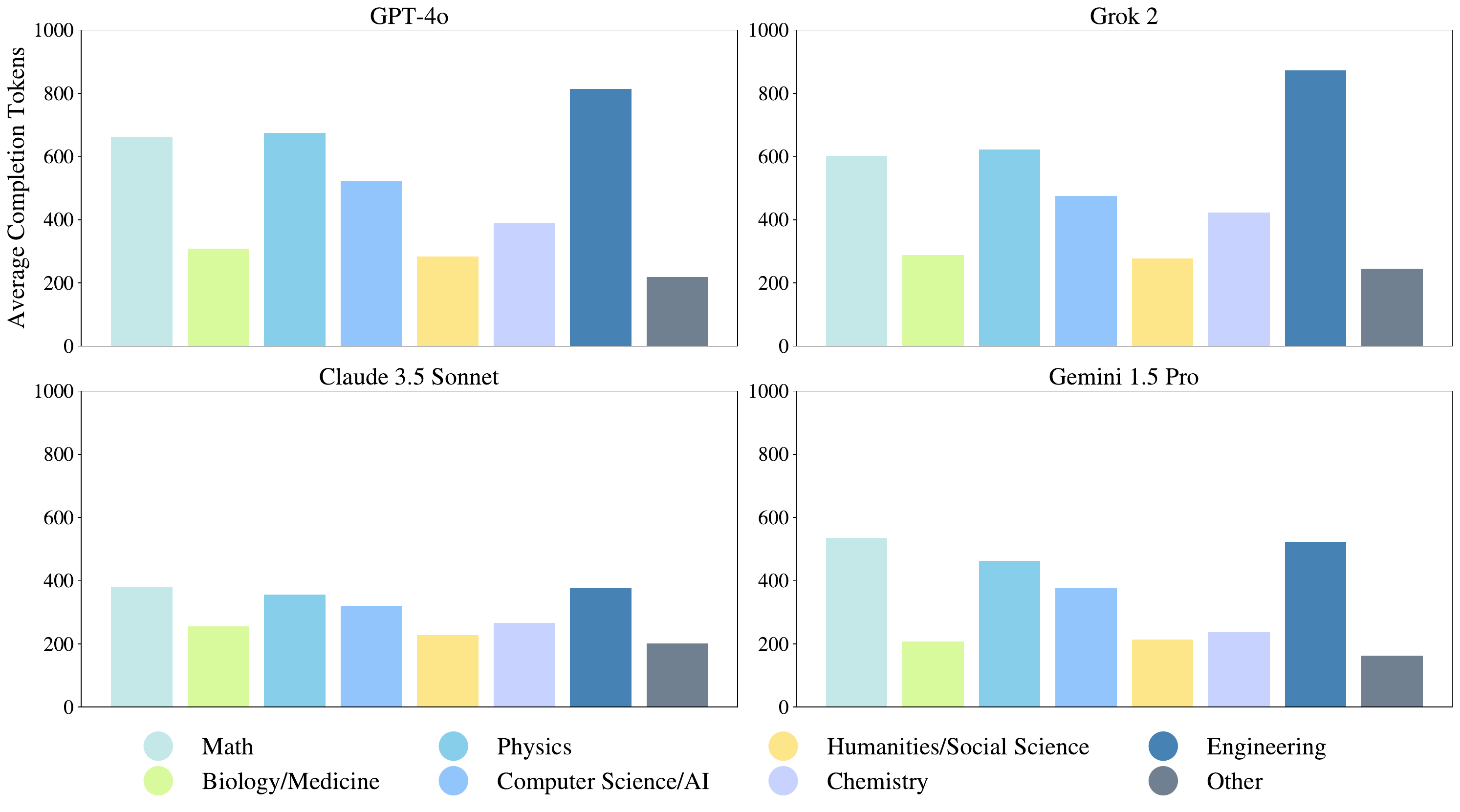}
    \caption{Average output token counts of non-reasoning models.}
    \label{fig:token-counts-nonreasoning}
\end{figure}

%% file: appendix/d-review_guidelines.tex
\subsection{Human Review Instructions}\label{app:dataset-human-review-instructions}
Questions which merely stump models are not necessarily high quality -- they could simply be adversarial to models without testing advanced knowledge. To resolve this, we employ two rounds of human review to ensure our dataset is thorough and sufficiently challenging as determined by human experts in their respective domains.

\subsubsection{Review Round 1} We recruit human subject expert reviewers to score, provide feedback, and iteratively refine all user submitted questions. This is similar to the peer review process in academic research, where reviewers give feedback to help question submitters create better questions. We train all reviewers on the instructions and rubric below.

\paragraph{Reviewer Instructions}
\begin{itemize}
    \item Questions should usually (but do not always need to) be at a graduate / PhD level or above. (Score 0 if the question is not complex enough and AI models can answer it correctly.)
    \begin{itemize}
        \item If the model is not able to answer correctly and the question is below a graduate level, the question can be acceptable.
    \end{itemize}
    
    \item Questions can be any field across STEM, law, history, psychology, philosophy, trivia, etc. as long as they are tough and interesting questions.
    \begin{itemize}
        \item For fields like psychology, philosophy, etc. we usually check if the rationale contains some reference to a book, paper or standard theories.
        \item For fields like law, the question text can be adjusted with ``as of 2024''. Make sure questions about law are time-bounded.
        \item Questions do not always need to be academic. A handful of movie, TV trivia, classics, history, art, or riddle questions in the dataset are OK.
        \item Trivia or complicated game strategy about chess, go, etc. are okay as long as they are difficult.
        \item We generally want things that require a high level of human intelligence to figure out.
    \end{itemize}
    
    \item Questions should ask for something precise and have an objectively correct, univocal answer.
    \begin{itemize}
        \item If there is some non-standard jargon for the topic/field, it needs to be explained.
        \item Questions must have answers that are known or solvable.
        \item Questions should not be subjective or have personal interpretation.
        \item Questions like ``Give a proof of\ldots''; ``Explain why\ldots''; ``Provide a theory that explains\ldots'' are usually bad because they are not closed-ended and we cannot evaluate them properly. (Score 0)
        \item No questions about morality or what is ethical/unethical. (Score 0)
    \end{itemize}
    
    \item Questions should be original and not derived from textbooks or Google. (Score 0 if searchable on web)
    
    \item Questions need to be in English. (Score 1 and ask for translation in the review if the question is written in a different language)
    
    \item Questions should be formatted properly. (Score 1-3 depending on degree of revisions needed)
    \begin{itemize}
        \item Question with numerical answers should have results approximated to max 2-3 decimals.
        \item Fix LaTeX formatting if possible. Models often get questions right after LaTeX formatting is added or improved.
        \item Questions that can be converted to text should be (converting images to text often helps models get them right).
    \end{itemize}
\end{itemize}

\paragraph{Other Tips}
\begin{itemize}
    \item Please write detailed justifications and feedback. This is going out to the question submitter so please use proper language and be respectful.
    \begin{itemize}
        \item Explanations should include at least some details or reference. If the rationale is unclear or not detailed, ask in the review to expand a bit.
        \item Please check if the answer makes sense as a possible response to the question, but if you do not have knowledge/context, or if it would take more than 5 minutes to solve, that is okay.
    \end{itemize}
    
    \item Please prioritize questions with no reviews and skip all questions with more than 3 reviews.
    
    \item Please double check that the model did actually answer the question wrong.
    \begin{itemize}
        \item Sometimes the exact match feature does not work well enough, and there are false negatives. We have to discard any exact match questions that a model got right.
    \end{itemize}
        
    \item On the \name{} dashboard, look at least 10 examples reviewed by the organizers before starting to review, and review the examples from training.
    
    \item The average time estimated to review a question 3-5 minutes.
    
    \item Use a ``-1 Unsure'' review if the person submitting seems suspicious or if you're not convinced their answer is right.
\end{itemize}

\renewcommand{\arraystretch}{1.2} 
\setlength{\tabcolsep}{5pt} 
\begin{tabular}{|l|l|p{8cm}|}
    \hline
    \textbf{Score} & \textbf{Scoring Guideline} & \textbf{Description} \\
    \hline
    \rowcolor{darkred}
    0 & Discard & The question is out of scope, not original, spam, or otherwise not good enough to be included in the HLE set and should be discarded. \\
    \hline
    \rowcolor{lightred}
    1 & Major Revisions Needed & Major revisions are needed for this question or the question is too easy and simple. \\
    \hline
    \rowcolor{lightorange}
    2 & Some Revisions Needed & Difficulty and expertise required to answer the question is borderline. Some revisions are needed for this question. \\
    \hline
    \rowcolor{lightyellow}
    3 & Okay & The question is sufficiently challenging but the knowledge required is not graduate-level nor complex. Minor revisions may be needed for this question. \\
    \hline
    \rowcolor{lightgreen}
    4 & Great & The knowledge required is at the graduate level or the question is sufficiently challenging. \\
    \hline
    \rowcolor{lightblue}
    5 & Top-Notch & Question is top-notch and perfect. \\
    \hline
    Unsure & - & Reviewer is unsure if the question fits the \name{} guidelines, or unsure if the answer is right. \\
    \hline
\end{tabular}

\subsubsection{Review Round 2} To thoroughly refine our dataset, we train a set of reviewers along with organizers to pick the best questions. These reviewers are identified by organizers from round 1 reviews as particularly high quality and thorough in their feedback. Different than the first round of reviews, reviewers are asked to grade both the question and look at feedback from round 1 reviewers. Organizers then approve questions based on reviewer feedback in this round. We employ a new rubric for this round below.

\renewcommand{\arraystretch}{1.2} 
\setlength{\tabcolsep}{5pt} 
\begin{tabular}{|l|l|p{8cm}|}
    \hline
    \textbf{Score} & \textbf{Scoring Guideline} & \textbf{Description} \\
    \hline
    \rowcolor{darkred}
    0 & Discard & The question is out of scope, not original, spam, or otherwise not good enough to be included in the HLE set and should be discarded. \\
    \hline
    \rowcolor{lightred}
    1 & Not sure & Major revisions are needed for this question or you’re just unsure about the question. Please put your thoughts in the comment box and an organizer will evaluate this.\\
    \hline
    \rowcolor{lightorange}
    2 & Pending & You believe there are still minor revisions that are needed on this question. Please put your thoughts in the comment box and an organizer will evaluate this. \\
    \hline
    \rowcolor{lightblue}
    3 & Easy questions models got wrong & These are very basic questions that models got correct or the question was easily found online. Any questions which are artificially difficult (large calculations needing a calculator, requires running/rendering code, etc.) should also belong in this category. The models we evaluate cannot access these tools, hence it creates an artificial difficulty bar. Important: “Found online” means via a simple search online. Research papers/journals/books are fine\\
    \hline
    \rowcolor{lightgreen}
    4 & Borderline & The question is not interesting OR The question is sufficiently challenging, but 1 or more of the models got the answer correct.\\
    \hline
    \rowcolor{lightpurple}
    5 & Okay to include in HLE benchmark & Very good questions (usually has score of 3 in the previous review round). You believe it should be included in the HLE Benchmark. \\
    \hline
    \rowcolor{lightyellow}
    6 & Top question in its category & Great question (usually has a score of 4-5 in the previous review round), at a graduate or research level. Please note that “graduate level” is less strict for Non-STEM questions. For Non-STEM questions and Trivia, they are fine as long as they are challenging and interesting. \\
    \hline

    \hline
\end{tabular}